\documentclass{article}

\usepackage{microtype}
\usepackage{graphicx}
\usepackage{subfigure}
\usepackage{booktabs} 

\usepackage{hyperref}

\usepackage[accepted]{icml2024}

\usepackage{amsmath}
\usepackage{amssymb}
\usepackage{mathtools}
\usepackage{amsthm}

\usepackage[capitalize,noabbrev]{cleveref}

\theoremstyle{plain}

\theoremstyle{definition}

\theoremstyle{remark}

\usepackage{rotating}
\usepackage{colortbl}
\usepackage{multirow}
\usepackage[textsize=tiny]{todonotes}

\usepackage{thmtools}
\usepackage{thm-restate}
\usepackage{xcolor}

\newcommand{\Real}{\mathbb{R}}

\newcommand{\Prob}[2]{\mathbb{P}_{#1}\left( #2 \right)}

\newcommand{\roundbrack}[1]{\left( #1 \right)}
\newcommand{\curlybrack}[1]{\left\lbrace #1 \right\rbrace}
\newcommand{\abs}[1]{\left| #1 \right|}
\newcommand{\indicator}[1]{\mathbf{1}\left\lbrace #1 \right\rbrace}

\icmltitlerunning{Adaptive Conformal Inference by Betting}

\begin{document}

\twocolumn[
\icmltitle{Adaptive Conformal Inference by Betting}




\begin{icmlauthorlist}
\icmlauthor{Aleksandr Podkopaev}{comp}
\icmlauthor{Darren Xu}{comp}
\icmlauthor{Kuang-chih Lee}{comp}
\end{icmlauthorlist}

\icmlaffiliation{comp}{Walmart Global Tech}

\icmlcorrespondingauthor{Aleksandr Podkopaev}{sasha.podkopaev@walmart.com}

\vskip 0.3in
]



\printAffiliationsAndNotice{}  

\begin{abstract}
Conformal prediction is a valuable tool for quantifying predictive uncertainty of machine learning models. However, its applicability relies on the assumption of data exchangeability, a condition which is often not met in real-world scenarios. In this paper, we consider the problem of adaptive conformal inference without any assumptions about the data generating process. Existing approaches for adaptive conformal inference are based on optimizing the pinball loss using variants of online gradient descent. A notable shortcoming of such approaches is in their explicit dependence on and sensitivity to the choice of the learning rates. In this paper, we propose a different approach for adaptive conformal inference that leverages parameter-free online convex optimization techniques. We prove that our method controls long-term miscoverage frequency at a nominal level and demonstrate its convincing empirical performance without any need of performing cumbersome parameter tuning.
\end{abstract}


\section{Introduction}

Accurate uncertainty estimation plays a crucial role in the practical deployment of machine learning models, particularly in contexts where model outputs impact downstream decision-making. A popular approach for quantifying predictive uncertainty is by using prediction sets: intervals in regression tasks or collection of labels in classification problems. The primary objective of such sets is to achieve valid coverage, meaning that they should cover the true labels with high probability (e.g., 90\%). In addition to coverage, the sharpness, or size of such prediction sets is extremely important in real-world applications. Conformal prediction~\citep{vovk2005alg} stands out as a versatile framework which is well-suited for this task: it allows for the construction of uncertainty quantification wrappers that can be seamlessly placed on top of arbitrary prediction models. 

Suppose that a model $\hat{f}:\Real^d \to \Real$ has been trained to generate real-valued predictions. One option is to resort to conformal predictors that output sets of the following form: $\hat{C}(x;s):= [\hat{f}(x)-s,\hat{f}(x)+s]$. Here, candidate prediction sets are parameterized by a single univariate parameter, denoted by $s$. The goal is to calibrate this parameter, i.e, determine a suitable $\hat{s}$ that ensures coverage. In this context, we remind the reader of a well-known technique of split conformal prediction, which relies on a holdout set not used during training: $\curlybrack{(X_i,Y_i)}_{i=1}^n$. Estimating errors via the absolute residuals, or the nonconformity scores: $R_i=|\hat{f}(X_i)- Y_i|$, and selecting $\hat{s}$ as the $\lceil(1-\alpha)(n+1)\rceil$-smallest value amongst $\{R_1,\dots, R_n, +\infty\}$ results in a conformal predictor that satisfies:
\begin{equation*}
    \Prob{}{Y_{\mathrm{test}}\in \hat{C}(X_{\mathrm{test}};\hat{s})}\geq 1-\alpha,
\end{equation*}
as long as $(X_1,Y_1)$, $\dots$, $(X_n,Y_n)$, $(X_{\mathrm{test}},Y_{\mathrm{test}})$ are exchangeable. We note that conformal inference is not specific to the above setting and has been extended in various directions. First, it offers much higher flexibility beyond the standard point forecasting model described above, e.g., it can be applied to recalibrate the prediction intervals associated with conditional quantile regression models~\citep{romano2019cqr}. Moreover, procedures that extend the split conformal framework beyond a single sample split, and hence allow better utilization of available data, have been developed, including procedures like Jackknife+~\citep{barber2021jackknife}. For a detailed overview of recent advancements and trends in the field of conformal inference, we refer the reader to~\citet{angelopoulos2023cp_intro}.

While conformal inference relies on the assumption of data exchangeability, it often fails to be met in practice. Examples include cases where data arrive sequentially over time, possibly exhibiting shifts in distribution, or where one is dealing with time series data. Despite the imposed practical challenges, there remains a huge demand for supplementing point predictions with \emph{valid} measures of uncertainty. In this work, we consider the problem of online conformal inference --- without imposing any distributional or dependency assumptions on the data generating process --- and focus on approaches that are applicable to arbitrary data streams.

We assume that the data are observed as a stream: $((X_t, Y_t))_{t\geq 1}$. At each time point $t$, the goal is to construct a prediction set for $Y_t$ using all of the previously observed data $\curlybrack{(X_i, Y_i)}_{i\leq t-1}$, as well as feature vector $X_t$. For brevity, we often capture the dependence of a conformal predictor on all of the available information that can be used at any time point (e.g., exogenous features or lagged response variables) using the index variable. Specifically, we write $\hat{C}_t(s):= \hat{C}_t(\curlybrack{(X_i, Y_i)}_{i\leq t-1}, X_t;s)$ to represent the prediction set for the response variable $Y_t$. Our objective is to design a conformal predictor whose observed long-term miscoverage rate is equal to the nominal level denoted as $\alpha$. In other words, we aim to construct a sequence of radii $(s_t)_{t\geq 1}$ so that the corresponding prediction sets satisfy:
\begin{equation}\label{eq:emp_coverage}
    \lim_{T\to\infty}\abs{\frac{1}{T}\sum_{t=1}^T \indicator{Y_t \notin \hat{C}_t(s_t)} - \alpha} = 0.
\end{equation}
In addition to assessing long-term coverage, we consider other performance metrics that are helpful in differentiating meaningful conformal predictors from trivial ones. For example, assuming that the response variables are bounded (i.e., $\abs{Y_t}\leq B$, $t\geq 1$, for some absolute constant $B>0$), a conformal predictor that switches at random between generating empty sets ($\alpha$ fraction of the time) and intervals $[-B,B]$ (the remaining $(1-\alpha)$ fraction of the time) technically satisfies~\eqref{eq:emp_coverage}, yet represents a practically useless tool for uncertainty quantification. To address this, we use the concept of regret, quantified as the cumulative pinball loss (since we are working with online quantile estimation) of a sequence of radii obtained using our method in comparison to an unknown benchmark point, as an additional metric. In particular, sub-linear regret bounds allow to justify the effectiveness of an adaptive conformal predictor in a more meaningful way compared to the coverage guarantee alone.

\paragraph{Related Work.} One of the earliest works where online convex optimization techniques have been applied in the context of uncertainty quantification is the one by~\citet{gibbs21aci}. Their methodology for learning a sequence of radii is based on applying online (sub)gradient descent to optimize the pinball loss. However, this approach has some limitations, including the need to specify the learning rate in advance and the potential for outputting empty or infinite prediction sets. Subsequent works by~\citet{zaffran2022aci, gibbs21aci_next, bhatnagar2023saocp, angelopoulos2023conformal} have introduced extensions of the above method to address some of those shortcomings.

The primary drawback of the aforementioned methods lies in their explicit dependence on a learning rate (or a specified grid thereof for approaches that utilize meta-learning to improve upon using a single learning rate), with the performance often being highly sensitive to such design choices. For example, higher learning rates promote adaptability to dynamic environments but may often lead to highly volatile prediction sets. The resulting online conformal predictors may oscillate between outputting overly small (anti-conservative) and excessively large (conservative) prediction sets for consecutive time steps, while still demonstrating empirical coverage close to the target level. Conversely, lower learning rates often result in conformal predictors that may sacrifice coverage for stability, potentially taking much longer to adapt to changes in distribution, hence failing to accurately represent uncertainty. Moreover, the selection of learning rates is heavily influenced by the scale of the errors, or the nonconformity scores, particularly in the case of real-valued responses. Heuristic approaches for approximating the scale, such as using a maximum or a high quantile of the historical response values, come with risks that may compromise the performance in practice (beyond the coverage guarantee being lost). These considerations introduce (potentially unnecessary) complexity if one aims to automate the implementation of the uncertainty quantification block in practice and become even more pronounced if uncertainty estimates are constructed for (a) a large collections of input data streams instead of a single one, and (b) multi-horizon forecasts rather than the one-step-ahead ones.

We note that several of the aforementioned works~\citep{bhatnagar2023saocp,gibbs21aci_next} proposed to supplement the coverage guarantee~\eqref{eq:emp_coverage} with regret guarantees that are stronger compared to the one considered in the current work. This is achieved via meta-learning: selected base models, e.g., the adaptive conformal predictors proposed by~\citet{gibbs21aci} with different learning rates, are subsequently aggregated using a meta-procedure. Our focus, however, is different as we consider practical algorithms designed to address the issues of cumbersome parameter tuning or even the necessity of selecting a grid of learning rates. We achieve this by leveraging parameter-free online convex optimization techniques with sub-linear regret bounds, particularly those that are based on coin betting~\citep{orabona2016coin_bet, cutkosky2018blackbox}.

Amongst other related works on conformal prediction with non-exchangeable data, we highlight methods that leverage reweighting schemes~\citep{tibs2019covariate, podkopaev2021label,lei2021ite, fannjiang2022feedback,candes2021surv_analysis} and approaches designed to handle time series data~\citep{chernozhukov18conf_inf,xu21conf_ts, stankeviciute2021conformal_ts, xu2023sequential_conformal}. We note that these methods either place some distributional assumptions (e.g., relationship between the source and the target domains for covariate/label shift, mixing assumptions for time series data) or characterize the coverage gap rather than guaranteeing coverage of the resulting predictor at a user-specified level.

\paragraph{Contributions.} In this work, we apply parameter-free optimization techniques to the problem of online conformal inference. We prove that the resulting conformal predictor controls the miscoverage rate at a pre-specified level. Through extensive simulations with focus on adaptability to distribution shifts, we demonstrate the compelling empirical performance of the proposed methods. Our approach nicely complements the existing methods in the literature due to its ease of implementation, computational efficiency, and absence of any parameter tuning.

\section{Betting-based Adaptive Conformal Inference}\label{sec:methodology}

We focus on conformal predictors that output sets of the following form: $\hat{C}_t(s):= [\hat{Y}_t-s,\hat{Y}_t+s]$, where the prediction $\hat{Y}_t$ is based on all information available prior to the true response $Y_t$ being revealed. We note that our methodology is applicable beyond such setting, e.g., it can be used to recalibrate the prediction intervals based on conditional quantile regression models: $\hat{C}_t(s):= [\hat{q}_t^{(\alpha/2)}-s;\hat{q}_t^{(1-\alpha/2)}+s]$, but we avoid the details for brevity. We refer the reader to~\citet{gupta2022nested} for more versions of the nested prediction sets for which our methodology is applicable. Let $S_t$ denote the radius of a smallest prediction set that contains the true response $Y_t$:
\begin{equation*}
\begin{aligned}
    S_t =& \inf\curlybrack{s\in\Real: Y_t\in \hat{C}_t(s)} \\
    =& \inf\curlybrack{s\in\Real: Y_t\in [\hat{Y}_t- s,\hat{Y}_t+s]} \\
    =& \abs{Y_t-\hat{Y}_t}.
\end{aligned}
\end{equation*}
Since the coverage event: $\{Y_t\in \hat{C}_t(s)\}$, is equivalent to $\curlybrack{S_t\leq s}$, the target property~\eqref{eq:emp_coverage} of the miscoverage being equal to the nominal level $\alpha$ can be expressed as:
\begin{equation}\label{eq:cov_eq_rep}
    \lim_{T\to\infty}\abs{\frac{1}{T}\sum_{t=1}^T \indicator{S_t\leq s}-(1-\alpha)} = 0.
\end{equation}
Hence, we can frame the task of constructing adaptive conformal predictors as a problem of learning $(1-\alpha)$-quantile of the nonconformity scores: $(S_t)_{t\geq 1}$, in an online fashion.

\paragraph{Quantile Estimation and Adaptive Conformal Inference.} Learning the quantiles of a distribution is achieved by optimizing the pinball loss, defined for $\beta$-quantile as
\begin{equation*}
\begin{aligned}
    \ell_\beta(s, S_t) =& \max \curlybrack{\beta(S_t-s), (1-\beta)(s-S_t)} \\
    =& \roundbrack{\indicator{s\geq S_t}-\beta}\roundbrack{s-S_t}.
\end{aligned}
\end{equation*}
The pinball loss is a convex and $\max\{\beta,1-\beta\}$-Lipschitz (in the first argument) loss function, whose subdifferential is given by
\begin{equation}\label{eq:pinball_subgrad}
    \partial \ell_\beta(s, S_t) = \begin{cases}
        \indicator{S_t\leq s}-\beta, & s\neq S_t, \\
        [-\beta, 1-\beta], & s = S_t.
    \end{cases}
\end{equation}
Taking $\beta=1-\alpha$, we recall that the updates corresponding to the online subgradient descent (OGD) take form:
\begin{equation}\label{eq:ogd_update}
\begin{aligned}
    s_{t+1} &= s_t - \eta \cdot\roundbrack{\indicator{S_t\leq s_t}-(1-\alpha)}\\
    &= s_t - \eta \cdot\roundbrack{\indicator{Y_t \in \hat{C}_t(s_t)}-(1-\alpha)}\\
    &= s_t - \eta \cdot\roundbrack{\alpha - \indicator{Y_t \notin \hat{C}_t(s_t)}}.
\end{aligned}
\end{equation}
Throughout this work, we let $g_t \in \partial \ell_{1-\alpha}(s, S_t)|_{s=s_t}$ denote the subgradients of the quantile losses at time steps $t=1,2,\dots$ The online gradient descent updates stated above admit a natural interpretation as an adjustment of the prediction interval's radius for the subsequent round in response to whether a conformal predictor covers the truth at a given time step: the radius is increased if a conformal predictor fails to cover the truth, and decreased otherwise. We note that in~\citet{gibbs21aci}, the authors did not directly apply the online subgradient descent to update the radii as stated in~\eqref{eq:ogd_update}. Instead, they applied it to update a sequence of quantile levels: $(\alpha_t)_{t\geq 1}$, and the radii were determined by computing the empirical quantiles of the residuals, or the nonconformity scores, at the corresponding levels. However, this approach faces the following issue: whenever some $\alpha_t$ falls outside the unit interval, the resulting conformal predictor outputs either infinite ($\alpha_t>1$) or empty ($\alpha_t<0$) prediction sets. While this problem does not arise when the radii are updated as per~\eqref{eq:ogd_update}, the scale of the radii becomes a crucial factor in determining an appropriate learning rate $\eta$ since the subgradients of the pinball loss are less than $\max\curlybrack{1-\alpha,\alpha}\leq 1$ in absolute value. 

We also consider a closely-related method for adaptive conformal inference that has been proposed by~\citet{bhatnagar2023saocp}. This approach builds upon scale-free online gradient descent (SF-OGD) introduced in~\citet{orabona2018sfonline}. The corresponding update rule takes form:
\begin{equation}\label{eq:sfogd_update}
    s_{t+1} = s_t - \eta \cdot \frac{\alpha - \indicator{Y_t \notin \hat{C}_t(s_t)}}{\sqrt{\sum_{i=1}^t \roundbrack{\alpha - \indicator{Y_i \notin \hat{C}_i(s_i)}}^2}},
\end{equation}
where, in contrast to~\eqref{eq:ogd_update}, the effective learning rate decays over time, being inversely proportional to the square root of the sum of the squared gradients. We note that SF-OGD still requires pre-specifying the learning rate, just as the standard OGD, and hence, requires considering the scale of the nonconformity scores. In both cases, the tuning process becomes much more complex if one applies adaptive conformal inference for multi-step forecasts or for a potentially large collection of input data streams. 

\paragraph{Our Approach.} We address the issues related to tuning the learning rates when learning conformal predictors by adopting parameter-free online convex optimization techniques. Specifically, we utilize optimization techniques that are based on coin betting~\citep{orabona2016coin_bet, cutkosky2018blackbox}. The high-level idea involves framing the learning process as a game where a gambler repeatedly places bets on the outcomes of continuous coin flips.

Let $W_t$ denote the gambler’s wealth at the end of round $t$. Starting with initial capital $W_0=1$, the gambler bets on the outcome of a coin flip $c_t \in [-1, 1]$ at each round $t$. The gambler is allowed to bet any amount $s_t$ on either heads or tails but is restricted from borrowing any money, i.e., we can write $s_t = \lambda_t W_{t-1}$ for some $\lambda_t\in [-1,1]$. The sign of $s_t$ specifies the gambler's choice between heads or tails (in general, $s_t$ may be negative) and the absolute value represents the corresponding betting amount. 

In the $t$-th round, the gambler gains $s_t c_t$ if $\mathrm{sign}(s_t) = \mathrm{sign}(c_t)$ and incurs a loss of $s_tc_t$ otherwise. Thus, we have: $W_t = W_{t-1} +s_tc_t = 1+\sum_{i=1}^ts_ic_i$. For the setting of i.i.d.\ coin flips ($c_t\in \{-1,+1\}$ are generated i.i.d.\ with a known probability of heads $p\in [0,1]$), the optimal strategy has been proposed by~\citet{kelly1956newinter}: he showed that betting $s_t = 2p-1$ yields more wealth than betting any other fixed fraction in the long run. For a sequence of possibly adversarial coin flips, \citet{kt1981perf} proposed a practical betting scheme that guarantees almost the same wealth as one could obtain betting any fixed fraction of wealth at each round. Moreover, the corresponding guarantee is known to be optimal up some constant factors~\citep{cb_lugosi2006}.

In our scenario, the coin outcomes are determined by the negation of the subgradients of the pinball loss defined in~\eqref{eq:pinball_subgrad}: $c_t = -g_t$ for $t\geq 1$. We consider two popular betting strategies: one based on Krichevsky-Trofimov (KT) estimator~\citep{kt1981perf}, extended to the case of continuous coins (i.e., $c_t\in [-1,1]$ instead of $\{-1,+1\}$) in~\citet{orabona2016coin_bet}, and a simple optimization procedure based on the Online Newton Step (ONS) method~\citep{hazan2007logarithmic,cutkosky2018blackbox}. In online learning, a standard performance metric is regret, which measures the cumulative loss of $(s_t)_{t=1}^T$ relative to an unknown benchmark point, denoted by $s^\circ$:
\begin{equation*}
    R_T(s^\circ) = \sum_{t=1}^T \ell_t(s_t)-\ell_t(s^\circ).
\end{equation*}
Betting games are useful for designing online convex optimization algorithms since the bounds on the minimum wealth can be used to derive the corresponding regret bounds. Both of the considered betting strategies yield online convex optimization algorithms with sub-linear regret in our setting. In particular, taking $\ell_t(s) = \ell_{1-\alpha}(s,S_t)$ for $t\geq 1$, we get a sequence of convex and $(1-\alpha)$-Lipschitz (assuming $\alpha<1/2$) loss functions. Therefore, for a sequence of radii $(s_t)_{t=1}^T$ obtained using the KT estimator it holds that:
\begin{equation}\label{eq:regret}
    R_T(s^\circ) \leq 1+ \abs{s^\circ} \sqrt{4T\ln\roundbrack{1+\abs{C T s^\circ}}}, \quad \forall s^\circ \in\Real,
\end{equation}
for some universal constant $C$~\citep{orabona2016coin_bet}. For online subgradient descent, the corresponding regret bound involves the learning rate parameter. More precisely, it can be shown for online subgradient descent that:
\begin{equation*}
    R_T(s^\circ) \leq \frac{(s^\circ)^2}{2\eta}+\frac{\eta T}{2}, \quad \forall s^\circ \in\Real,
\end{equation*}
and hence, the learning rate which minimizes the upper bound is $\eta = \abs{s^\circ}/\sqrt{T}$. As discussed later in this Section, the boundedness of the nonconformity scores: $\abs{S_t}\leq D$ for some $D>0$ and all $t\geq 1$, is a necessary condition to ensure long-term coverage~\eqref{eq:emp_coverage}. In this case, using $\eta = D/\sqrt{T}$ results in regret bound: $D\sqrt{T}$, which is known to be optimal for bounded domains up to multiplicative constants. However, implementing the resulting algortihm in practice still requires an explicit knowledge of $D$. In contrast, utilizing KT betting results in a sub-optimal regret bound up to logarithmic factors (which is a secondary metric of our interest), but allows to avoid tuning any parameters.

We summarize the adaptive conformal predictor that utilizes KT betting strategy in Algorithm~\ref{alg:adapt_conf_kt} and defer the description of the adaptive conformal predictor that uses the ONS betting scheme to Appendix~\ref{appsec:omitted}.

\begin{algorithm}
\caption{KT-based Adaptive Conformal Predictor.}
\label{alg:adapt_conf_kt}
\begin{algorithmic}
\STATE \textbf{Initialize:} $\alpha\in(0,1)$, $W_0=1$, $\lambda_1=0$, $s_1=0$.
\FOR{$t=1,2,\dots$}
\STATE Produce a forecast $\hat{Y}_{t} = f_t(X_t, \curlybrack{(X_i,Y_i)}_{i\leq t-1})$ and output a set: $\hat{C}_t(s_t)= [\hat{Y}_{t}-s_t;\hat{Y}_{t}+s_t]$;
\STATE Observe $Y_t$ and compute error: $S_t = \abs{Y_t-\hat{Y}_t}$;
\STATE Compute $g_t \in \partial \ell_{1-\alpha}(s, S_t)|_{s=s_t}$ as per~\eqref{eq:pinball_subgrad};
\STATE Set $W_t = W_{t-1}-g_t s_t$;
\STATE Set $\lambda_{t+1} = \frac{t}{t+1}\lambda_t - \frac{1}{t+1} g_t$;
\STATE Set $s_{t+1}=\lambda_{t+1}W_{t}$;
\ENDFOR
\end{algorithmic}
\end{algorithm}

While the sub-linear regret guarantees are helpful in eliminating trivial conformal predictors, the coverage guarantee outlined in~\eqref{eq:emp_coverage} does not directly follow from the regret bound. These guarantees have to be derived independently. In the following result, we establish that the proposed approach for online conformal inference based on KT betting strategy attains a long-term miscoverage rate precisely equal to the nominal level $\alpha$. The proof is deferred to Appendix~\ref{appsec:proofs}.

\begin{restatable}[]{theorem}{miscoverage}\label{thm:goldbach}
Fix the target miscoverage level $\alpha \in (0,1/2)$. Suppose that the nonconformity scores are bounded: $S_t\in [0, D]$ for $t=1,2,\dots$, for some $D>0$. Then the adaptive conformal predictor defined in Algorithm~\ref{alg:adapt_conf_kt} satisfies the long-term coverage guarantee~\eqref{eq:emp_coverage}.
\end{restatable}

The boundedness of the nonconformity scores is the only assumption which is made in Theorem~\ref{thm:goldbach} to ensure coverage of the proposed conformal predictor. The method itself does not depend on the explicit knowledge of such bound. It is easy to see that for the KT-based online conformal predictor outlined in Algorithm~\ref{alg:adapt_conf_kt} the assumption regarding bounded scores is indeed necessary for achieving~\eqref{eq:emp_coverage}. If this assumption is violated, then one can easily construct an adversarial example where miscoverage rate is actually equal to one: once the radius $s_t$ is predicted, it is always possible to choose a response value $Y_t$ that lies outside of the predicted interval, resulting in error at each round. Finally, we note that the same argument can be used to show that the boundedness assumption is also necessary for conformal predictors whose radii are updated according to~\eqref{eq:ogd_update} or~\eqref{eq:sfogd_update} to satisfy the long-term coverage guarantee~\eqref{eq:emp_coverage}.

\section{Experiments}\label{sec:exps}

In our simulation study, we consider a collection of simulated and real datasets where the data distribution changes over time, and hence, the exchangeability assumption is violated. Throughout all experiments, we fix the target coverage level at 90\% ($\alpha=0.1$). We compare adaptive conformal predictors learned using the proposed betting scheme against those that are learned using OGD~\eqref{eq:ogd_update} and SF-OGD~\eqref{eq:sfogd_update}. We demonstrate that our method --- without performing any parameter tuning --- achieves performance that either matches or is close to that of a conformal predictor obtained by deploying versions of online gradient descent with carefully tuned learning rates. 

\begin{figure}[!ht]
    \centering
    \includegraphics[width=0.9\linewidth]{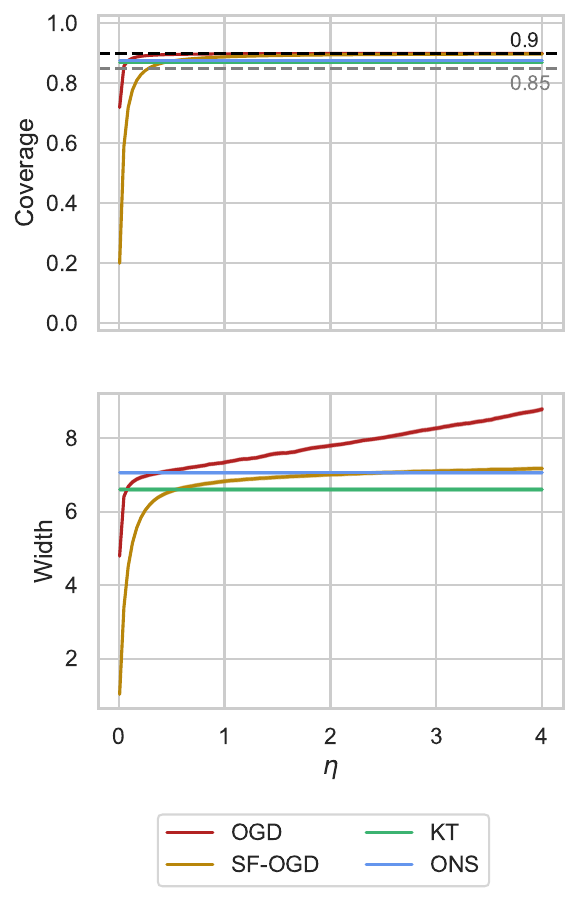}
    \caption{Comparison of the proposed conformal predictor against those learned via OGD/SF-OGD with different learning rates. The performance of the conformal predictors learned via OGD/SF-OGD is sensitive to the choice of the learning rate, whereas the performance of the betting-based ones is close to (in terms of coverage and width) to that of the carefully tuned alternative methods. The results are aggregated over 200 random seeds.}
    \label{fig:changepoint_summary}
\end{figure}

\paragraph{Changepoint Setting.} Following~\citet{bacarati2023beyond}, we consider a changepoint setting setting where the data $\{(X_t,Y_t)\}_{t=1}^n$ are generated according to a linear model: $Y_t = X_t^\top \beta_t+\varepsilon_t$, $X_t\sim \mathcal{N}(0, I_4)$, $\varepsilon_t\sim \mathcal{N}(0,1)$, $t\geq 1$. We consider the following scenario:
\begin{equation*}
\begin{aligned}
    \beta_t &= \beta^{(0)} = (2,1,0,0)^\top, \quad t=1,\dots,500,\\
    \beta_t &= \beta^{(1)} = (0,-2,-1,0)^\top, \quad t=501,\dots,1500,\\
    \beta_t &= \beta^{(2)} = (0,0,2,1)^\top, \quad t=1501,\dots,2000,
\end{aligned}
\end{equation*}
where two changes in the coefficients happen up to time 2000. For prediction, we first use a standard linear regression model whose coefficients are learned by optimizing the least squares objective on observed data prior to a given time step. In Figure~\ref{fig:changepoint_summary}, we compare our adaptive conformal predictors against those that are trained using variants of gradient descent with varying learning rates. 

The top plot shows that the empirical coverage of the conformal predictors learned via versions of online gradient descent is nearly equal to the nominal level whenever the learning rates are high enough. Although our conformal predictor demonstrates slightly lower coverage (around 88\%), such difference is typically of minor practical importance. When the learning rates are too small, the OGD-based conformal predictors demonstrate empirical coverage that is significantly lower than the target level. Conversely, with learning rates that are overly high, the average width of the output sets increases, resulting in overly conservative sets, as observed in the bottom plot. In fact, such conformal predictors oscillate between outputting overly narrow and overly wide sets (since the learning rate is high and fixed). 

\begin{figure*}[!ht]
    \centering
    \includegraphics[width=\linewidth]{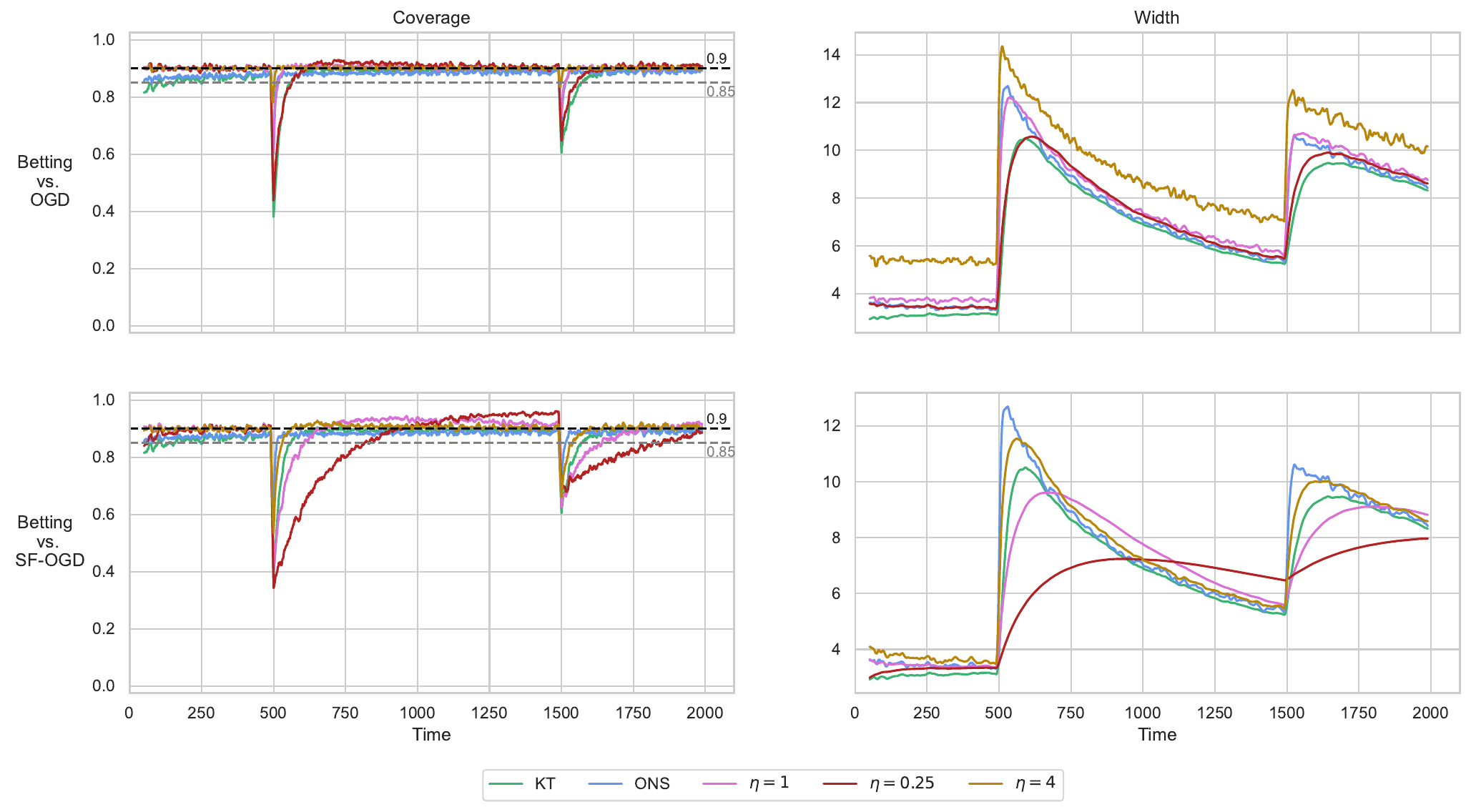}
    \caption{Comparison of the conformal predictor trained using parameter-free optimization techniques (KT, ONS) against those trained using variants of online gradient descent with varying learning rates (OGD, SF-OGD). We avoid plotting results observed for the first 50 observations. The results are aggregated over 250 random seeds and smoothed using rolling window of size 10.}
    \label{fig:exp_changepoint}
\end{figure*}

While useful, the above findings provide limited insights into the adaptability of conformal predictors to changes in distribution. To illustrate such adaptability properties, we compare the localized coverage and width of our conformal predictors and those trained via online gradient descent for three particular choices of learning rates in Figure~\ref{fig:exp_changepoint}. We observe a drastic impact of a learning rate on the performance of the resulting conformal predictor. For the OGD-based method, the output sets are generally overly conservative (top-right plot) for large learning rate ($\eta=4$). This issue is addressed by SF-OGD (bottom-right plot), whose learning rate effectively decreases over time. However, if the learning rate becomes too low ($\eta<1$), the localized coverage of SFOGD-based conformal predictor recovers very slowly after changes in distribution take place.

\begin{figure}[!ht]
    \centering
    \includegraphics[width=\linewidth]{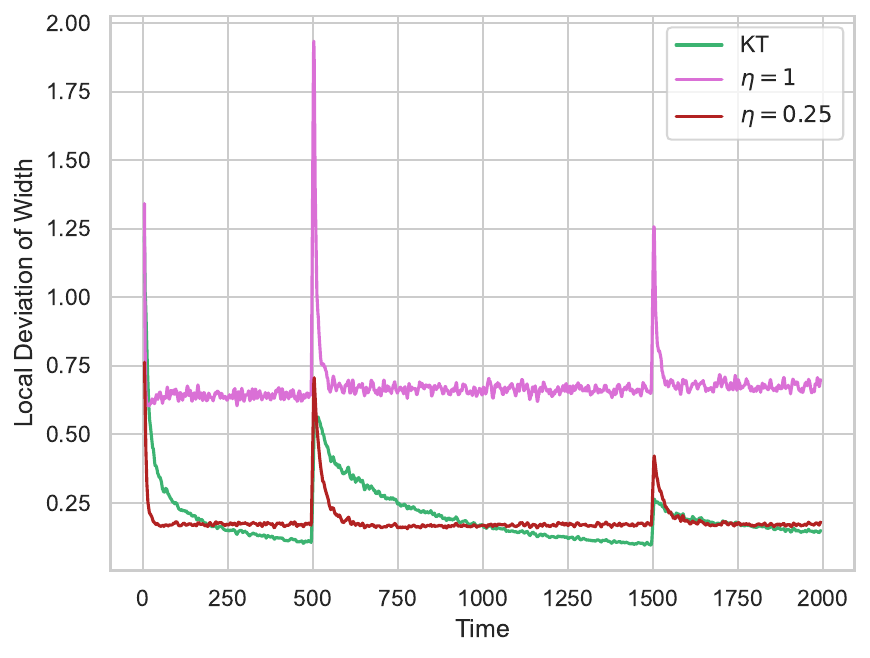}
    \caption{Local deviation of the width of the uncertainty sets returned by KT-based and OGD-based conformal predictors. If the learning rate for OGD-based conformal predictor is set too high, the width of the output sets may change abruptly between consecutive time steps. Deviations are computed using rolling window of size 10 and are averaged over 250 random seeds.}
    \label{fig:local_deviation}
\end{figure}

As alluded to before, if the learning rate for OGD-based conformal predictors is set too high, the resulting sets may become volatile. We illustrate this problem on Figure~\ref{fig:local_deviation} where the KT-based conformal predictor is compared against the one based on OGD with $\eta\in \{0.25, 1\}$. For each approach, we estimate local deviation of the interval width using a rolling window of size 10. Although the empirical coverage and average width of the uncertainty intervals produced by conformal predictor based on OGD with $\eta=1$ is close to that of the KT-based one (Figure~\ref{fig:exp_changepoint}), we observe that the local deviation of the interval width for the former method is much higher, indicating that the corresponding width changes abruptly between consecutive time steps.

Under an abrupt change in the data distribution, the predictive accuracy may drop if the model is trained on all data (without considering potential shifts in distribution). Predictive models which are trained using online gradient descent or utilize weighting schemes, with higher weights being assigned to the most recent datapoints, may adapt to shifts in distribution much faster. We consider a second option and refer the reader to Appendix~\ref{appsubsec:changepoint_weighted_ls} for a comparison between various methods for adaptive conformal inference when a linear model, whose coefficients are learned by optimizing the weighted least squares objective, is used.

\begin{figure*}[!ht]
    \centering
    \subfigure[One-step ahead.]{\includegraphics[width=0.45\linewidth]{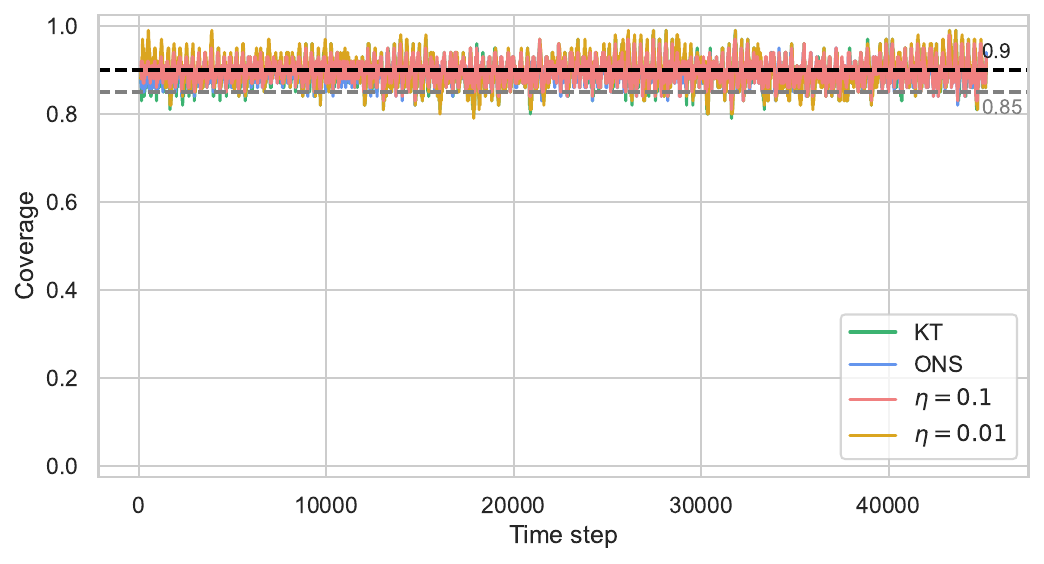}
    \label{subfig:one_step_cov}}
    \hspace{0.25in}
    \subfigure[One-step ahead.]{\includegraphics[width=0.45\linewidth]{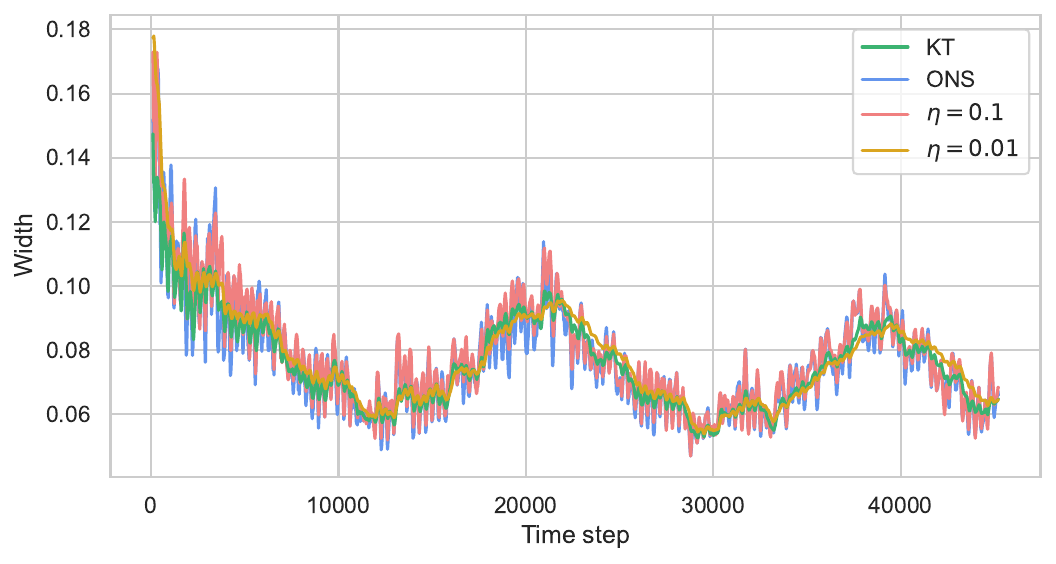}
    \label{subfig:one_step_width}}
    \caption{Comparison between conformal predictors constructed using betting schemes and SF-OGD~\eqref{eq:sfogd_update} with $\eta\in\{0.01,0.1\}$ for one-step ahead forecasting. For all methods, the empirical coverage is near the nominal level and the resulting prediction sets have roughly similar width. The results are smoothed over a rolling window of 100 observations. }
    \label{fig:electricity_sfogd}
\end{figure*}

\paragraph{Electricity Demand Data.} Next, we consider the dataset for forecasting the electricity demand in New South Wales~\citep{harries1999elect}. Following~\citet{angelopoulos2023conformal}, we use $\mathrm{AR}(3)$ model as an underlying predictor. In Figure~\ref{fig:electricity_sfogd}, we compare coverage and width respectively of conformal predictors constructed using betting schemes against those based on online gradient descent with varying learning rates. We present the results for SF-OGD only, deferring those for OGD to Appendix~\ref{appsubsec:electricity_demand}.

We observe that four conformal predictors demonstrate similar performance: for all methods, the empirical coverage is near the nominal level and the resulting prediction sets have roughly similar width. As we illustrate in Appendix~\ref{appsubsec:electricity_demand} (particularly, Figure~\ref{fig:kt_sfogd_ts}), the resulting prediction sets become visually indistinguishable after processing a relatively small number of observations. In contrast, OGD with the same learning rate ($\eta=0.1$) yields conformal predictors that are significantly wider on average than that of alternative methods. This is due to using a fixed learning rate throughout the whole process; see Figure~\ref{fig:electricity_histograms} for details.

In applications, practitioners are often interested in multi-step forecasting for some horizon $H$. Adaptive conformal predictors may handle such cases by simply associating each step with a separate radius: $s^{(1)}, \dots, s^{(H)}$, and deploying online optimization schemes (e.g., one in Algorithm~\ref{alg:adapt_conf_kt}) independently for each of the parameters. On the same electricity dataset, we consider the problem of uncertainty quantification in multi-step forecasting, setting the horizon $H=5$. We use $\mathrm{AR}(3)$ model and utilize a simple approach when $k$-step ahead forecast is used as an input feature for constructing a forecast on $(k+1)$-st step. The parameters of the model are updated each time the next 5 true responses are revealed. In Table~\ref{table:ts_multi_summary}, we summarize average empirical coverage and average size of the prediction set for conformal predictors learned using KT betting scheme and versions of online gradient descent with $\eta=0.01$. In terms of global metrics, all methods demonstrate similar performance.

\begin{figure*}[!ht]
    \centering
    \subfigure[Five-step ahead.]{\includegraphics[width=0.45\linewidth]{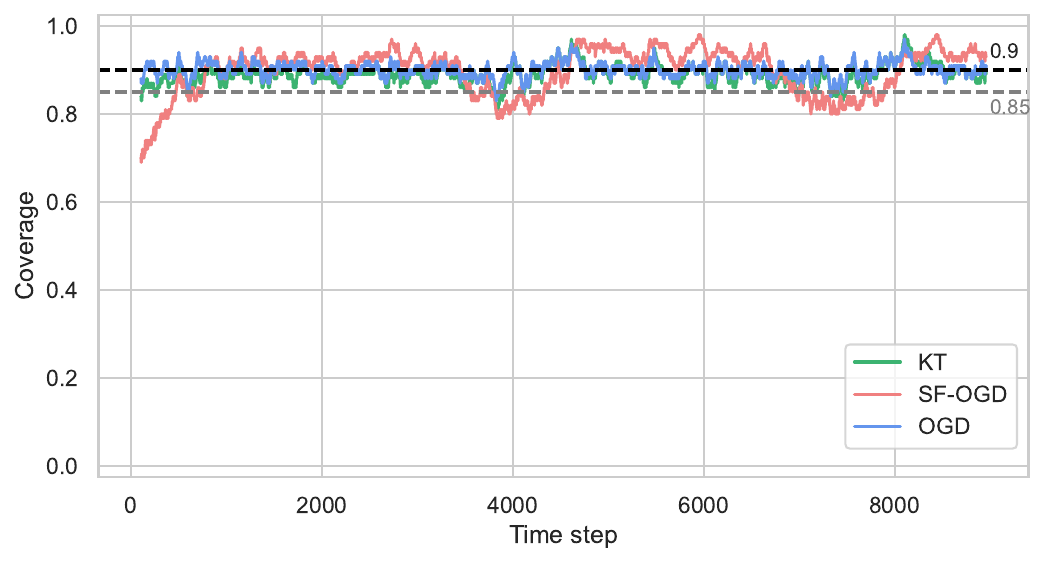}
    \label{subfig:five_step_cov}}
    \hspace{0.25in}
    \subfigure[Five-step ahead.]{\includegraphics[width=0.45\linewidth]{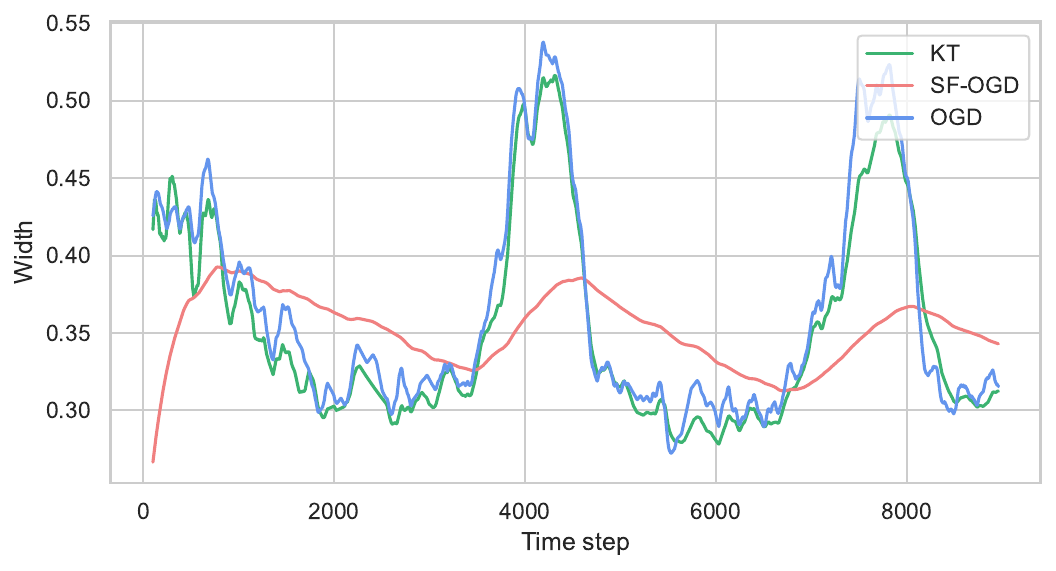}
    \label{subfig:five_step_width}}
    \caption{Comparison between conformal predictors constructed using betting schemes and versions of versions of online gradient descent~\eqref{eq:ogd_update} and~\eqref{eq:sfogd_update} with $\eta = 0.01$ for five-step ahead forecasting. The results are presented for the fifth step and are smoothed over a rolling window of 100 observations.}
    \label{fig:electricity_multihorizon}
\end{figure*}

In Figure~\ref{fig:electricity_multihorizon}, we illustrate localized coverage and width for all methods, restricting the attention to the last step in the forecasting horizon ($k=5$). In this case, we observe that the KT-based conformal predictor exhibits behavior that is closer to that based on OGD~\eqref{eq:ogd_update} rather than to that based on SF-OGD~\eqref{eq:sfogd_update}: for the two former methods, localized coverage is more tightly concentrated around the nominal level. In contrast, the conformal predictor based on SF-OGD happens to either undercover or be overly conservative over long periods of time, failing to respond quickly to the changes in the data distribution.

\begin{table}[!ht]
\centering
\begin{tabular}{c|c c c|c c c} 
 \multicolumn{4}{c}{\hspace{0.1in} Coverage} & \multicolumn{3}{c}{Width} \\
 \hline
 $k$ & KT& OGD& SF-OGD& KT& OGD& SF-OGD\\
\hline
1 & 89.1 & 89.9 & 90.1 & 7.58 & 8.36 & 7.92 \\
 2 & 89 & 89.9 & 90.1 & 14.3 & 14.9& 14.6 \\
 3 & 89 & 89.9 & 89.8 & 21.6 & 22.6 & 21.8 \\
 4 & 89 & 89.9 & 89.5 & 28.6 & 30 & 28.7 \\
 5 & 88.9 & 89.8 & 89.2 & 35.3& 36.3& 35 
\end{tabular}
    \caption{The results for uncertainty quantification in $k$-step ahead electricity demand forecasting. The empirical coverage of conformal predictors learned using versions of online gradient descent is closer to the nominal level, yet the coverage of that learned using betting schemes is only slightly below. The KT-based conformal predictor yields shorter prediction sets on average (four out of five cases). The empirical coverage is shown in percentages. The average width of prediction sets has been multiplied by 100.}
\label{table:ts_multi_summary}
\end{table}

\begin{figure}[!ht]
    \centering
    \includegraphics[width=\linewidth]{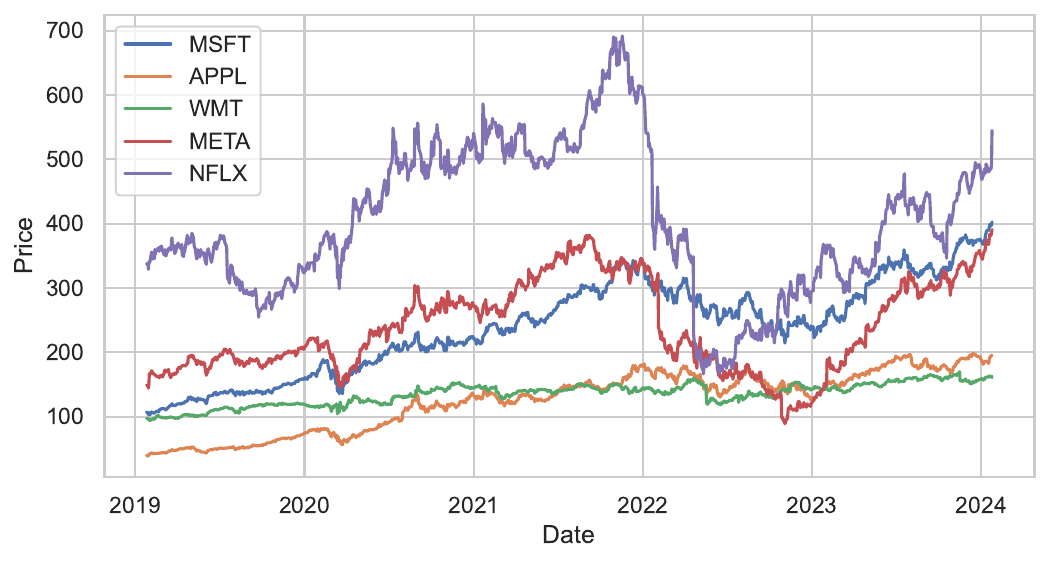}
    \caption{Visualization of the prices for the selected stocks over five years.}
    \label{fig:stock_prices_vis}
\end{figure}

\paragraph{Stock Prices Data.} Finally, we consider uncertainty quantification in the problem of forecasting stock prices. In particular, we use the closing prices of five different stocks: Apple (\texttt{APPL}), Meta (\texttt{META}), Microsoft (\texttt{MSFT}), Netflix (\texttt{NFLX}), and Walmart (\texttt{WMT}), collected over a five-year period (from January, 25th 2019 to Jan, 24th 2024); see Figure~\ref{fig:stock_prices_vis}\footnote{\href{https://www.nasdaq.com/market-activity/stocks/wmt/historical}{Example of the data source for one of the stocks.}}. Rather than forecasting the closing price for the next day only, we consider multi-horizon forecasting for each calendar week (i.e., forecasting horizon $H=5$): each day is associated with the corresponding radius which is updated on a weekly basis, i.e., $s^{(1)}$ is a radius that is used to construct uncertainty estimates for Mondays exclusively. We take into account the days of market closure due to holidays as follows. For example, if the  first trading day of a week happens to be Wednesday, we use the radius $s^{(3)}$ to construct the corresponding prediction interval. Subsequently, after prices in a given week are observed, there is no update to $s^{(1)}$ and $s^{(2)}$. 

We use the Prophet~\citep{taylor2018prophet} as our prediction model and operate in the log-space for forecasting. For each stock, the first 25 weeks of data are used to train the initial model, followed by retraining at the end of each subsequent week. For conformal predictors based on variants of online gradient descent, we compute the in-sample residuals for the initial model and use the empirical absolute error as a learning rate. Although the in-sample residuals underestimate the out-of-sample residuals, one may want to try a smaller learning rate, particularly for the OGD-based conformal predictor~\eqref{eq:ogd_update}. However, our empirical observations indicated that such choice worsens the results in terms of coverage. In addition to conformal methods, we also consider prediction intervals that are provided by Prophet as a native uncertainty quantification tool.

In Table~\ref{table:agg_stock}, we present an overview of the empirical coverage results aggregated across five stocks. We observe that all methods for uncertainty quantification demonstrate an average coverage lower than the nominal level. The uncertainty intervals provided by Prophet stand out as significantly sub-optimal compared to alternative methods. In Appendix~\ref{appsubsec:stocks}, we further illustrate Prophet fails to adapt to multi-step forecasting by only marginally increasing in set size relative to the number of steps ahead. Amongst other methods, conformal predictor which is based on OGD~\eqref{eq:ogd_update} demonstrates the coverage that is closest to the nominal level. However, we note that the KT-based conformal predictor is only slightly inferior, while being insensitive to parameter tuning. Unlike standard OGD, the effective learning rate of SF-OGD is decreasing over time. This in turn has a direct impact on the poor empirical coverage of the resulting conformal predictor: decreasing learning rate happens to hurt the ability of the uncertainty estimates to appropriately adjust in response to the drop of model accuracy. We refer the reader to Appendix~\ref{appsubsec:stocks} for additional results about the performance of different adaptive conformal predictors in stock price forecasting. In particular, we demonstrate that in some cases, the KT-based conformal predictor yields shorter intervals than that based on OGD, despite showing similar coverage.

\begin{table}[!ht]
\centering
\begin{tabular}{c|c c c c } 
 \multicolumn{5}{c}{\hspace{0.1in} Coverage}  \\
 \hline
 $k$ & KT& OGD& SF-OGD& Native method\\
\hline
1 &  84.8 & 86 & 79.6 & 69.4\\
 2 & 85.0 & 86.4 & 79.5 & 67.6 \\
 3 & 84.9 & 86.4 & 78.8 & 62.9 \\
 4 & 84.6 & 85.9& 76.9& 60.4 \\
 5 & 84.6 & 85.4 & 76.3 & 58.8
\end{tabular}
    \caption{Empirical coverage comparison in $k$-step ahead stock price forecasting. The OGD-based conformal predictor demonstrates coverage that is closest to the nominal level, with the KT-based one being slightly inferior. The native prediction intervals offered by Prophet show the lowest coverage across all steps.}
\label{table:agg_stock}
\end{table}

\section{Conclusion}

A number of methods have been recently proposed for online conformal inference, primarily utilizing versions of online gradient descent. Such methods are generally sensitive to the choice of (possibly a grid of) learning rates, and hence, usually require careful tuning. Our primary contribution lies in demonstrating that parameter-free online convex optimization techniques can effectively address this issue, resulting in a compelling method for adaptive conformal inference. Despite its simplicity, our approach is advantageous from several standpoints. First, our online conformal predictor provably achieves long-term coverage. Second, additional properties, such as sub-linear regret, justify its practical utility. We note that the absence of tuning comes at a cost: our method is guaranteed to achieve correct coverage rate in the limit, it may demonstrate coverage that is lower than the nominal level in finite-sample regime (although, the difference is usually small and is of little practical interest). Methods that are based on versions of online gradient descent often demonstrate marginal coverage that is closer to the the nominal level, but run into risk of failing to adapt to distribution shifts, yielding either overly conservative or overly optimistic prediction sets over spans of time as a result. Empirical evidence demonstrates that our method generally performs only slightly worse or matches the performance of methods based on gradient descent with carefully tuned parameters. Therefore, using betting-based conformal inference can be advantageous in scenarios where precision is prioritized (given an acceptable coverage). The primary strength of our method lies in its simplicity and ease of implementation, making it a practical and accessible choice for applications. Therefore, we view it as a useful method in a toolbox of machine learning practitioners. 

\section*{Impact Statement}

This paper presents general methodological work whose goal is to advance the field of Machine Learning. By providing open access to the code as a supplement for the purposes of transparency and reproducibility, our work aims to reach better understanding within the research community. There are many potential societal consequences of our work, none which we feel must be specifically highlighted here.

\bibliography{refs}
\bibliographystyle{icml2024}

\newpage
\appendix
\onecolumn
\noindent\rule{\textwidth}{2pt}
\begin{center}
    \vspace{7pt}
    {\Large Appendix}
\end{center}
\noindent\rule{\textwidth}{2pt}

\section{Omitted Details}\label{appsec:omitted}

\paragraph{Online Conformal Predictor with ONS Bets.} A complete description of online conformal predictor that uses bets provided by online Newton step (ONS) is provided in Algorithm~\ref{alg:adapt_conf_ons}.

\begin{algorithm}[!ht]
\caption{ONS-based Online Conformal Predictor.}
\label{alg:adapt_conf_ons}
\begin{algorithmic}
\STATE \textbf{Initialize:} $W_0=1$, $\lambda_1=0$, $A_0=1$, $\alpha\in(0,1)$.
\FOR{$t=1,2,\dots$}
\STATE Produce a forecast $\hat{Y}_{t} = f_t(X_t, \curlybrack{(X_i,Y_i)}_{i\leq t-1})$ and output a set: $\hat{C}_t(s_t)= [\hat{Y}_{t}-s_t;\hat{Y}_{t}+s_t]$;
\STATE Observe $Y_t$ and compute error: $S_t = \abs{Y_t-\hat{Y}_t}$;
\STATE Compute $g_t = \partial \ell_{1-\alpha}(s, S_t)|_{s=s_t}$;
\STATE Set $W_t = W_{t-1}-g_t s_t$;
\STATE Set $z_t=g_t/(1-\lambda_t g_t)$;
\STATE Set $A_t = A_{t-1}+ z_t^2$;
\STATE Set $\lambda_{t+1} = \roundbrack{\roundbrack{\lambda_t- \frac{2}{2-\log(3)} \frac{z_t}{A_t}}\vee -\frac{1}{2}}\wedge \frac{1}{2}$;
\STATE Set $s_{t+1}=\lambda_{t+1}W_{t}$;
\ENDFOR
\end{algorithmic}
\end{algorithm}

While ONS betting scheme tends to yield adaptive conformal predictors with very impressive empirical performance, the corresponding betting fractions: $(\lambda_t)_{t\geq 1}$, are defined recursively which complicates the theoretical analysis of the resulting adaptive conformal predictors.

\section{Proofs}\label{appsec:proofs}

\miscoverage*

\begin{proof}

\begin{enumerate}
\item First, note under the assumption that the nonconformity scores are bounded: $S_i\leq D$, $i=1,2,\dots$, for some $D>0$, the following statements hold:
\begin{enumerate}
    \item Suppose that for some $i\geq 1$, it happens that the \emph{predicted} radius $s_i$ exceeds the upper bound $D$: $s_i>D$. Since $s_i = \lambda_i\cdot W_{i-1}$ and the wealth is nonnegative $W_{i-1}\geq 0$, it implies that $\lambda_i>0$. Further, the corresponding (sub)gradient is $g_i = \alpha- \textbf{1}\{Y_i\notin \hat{C}_i(X_i)\} = \alpha - \textbf{1}\{S_i>s_i\} = \alpha$, which in turn implies that $W_i = W_{i-1}(1-\lambda_i g_i)<W_{i-1}$. For KT estimator, it holds that: $\lambda_{i+1}=\frac{i}{i+1}\lambda_i - \frac{1}{i+1}g_i<\lambda_i$. In other words, we get that $s_{i+1} = \lambda_{i+1}W_i<s_i$, meaning that the predicted radius for the next step necessarily decreases, and this process repeats until the predicted radius becomes less or equal than $D$. 
    \item Suppose that for some $i\geq 1$, it holds that: $s_i\geq 0$, but $s_{i+1}<0$. Then it has to be the case that $s_{i+2}>0$. Indeed, $s_i\geq 0$ implies that $\lambda_i\geq 0$ and $s_{i+1}<0$ implies that $\lambda_{i+1} < 0$. Next, note that for KT estimator, it holds that:
    \begin{equation*}
    0>\lambda_{i+1} = \frac{i}{i+1}\lambda_{i} - \frac{1}{i+1}g_{i},   
    \end{equation*}
    which implies that $g_i> 0$, and hence, $g_i = \alpha$. Since $S_{i+1}\geq 0$, it holds that $g_{i+1}  = \alpha - \textbf{1}\{S_{i+1}>s_{i+1}\} = \alpha-1$. Finally, 
    \begin{equation*}
    \begin{aligned}
        \lambda_{i+2} &= \frac{i+1}{i+2}\lambda_{i+1} - \frac{1}{i+2}g_{i+1}\\
        &=\frac{i+1}{i+2}\frac{i}{i+1}\lambda_{i} -\frac{i+1}{i+2}\frac{1}{i+1}g_i- \frac{1}{i+2}g_{i+1}\\
        &=\frac{i}{i+2}\lambda_{i} -\frac{1}{i+2}(g_i+g_{i+1}).
    \end{aligned}
    \end{equation*}
    Hence, since $\lambda_{i}\geq 0$ and $g_i+g_{i+1} = 2\alpha-1<0$ (where we make a mild assumption that $\alpha<0.5$), we conclude that $\lambda_{i+2}>0$, and hence, $s_{i+2}>0$.
    
    \end{enumerate}

\item Since for any $t\geq 1$, $W_t = 1-\sum_{i=1}^ts_ig_i\geq 0$, we get that $\sum_{i=1}^ts_ig_i\leq 1$. On the other hand, recall that if $s_i>D$, then we have that: $g_i=\alpha>0$, and if $s_i<0$, then $g_i=\alpha-1<0$. Hence,
\begin{equation*}
\begin{aligned}
    \sum_{i=1}^{t}g_i s_i &=  \sum_{i=1}^{t} \underbrace{g_i s_i}_{>0} \cdot \textbf{1}\curlybrack{s_i> D}+\sum_{i=1}^{t}g_i s_i \cdot \textbf{1}\curlybrack{s_i\in[0,D]}+\sum_{i=1}^{t} \underbrace{g_i s_i}_{>0} \cdot \textbf{1}\curlybrack{s_i< 0} \\
    &\geq \sum_{i=1}^{t}g_i s_i \cdot \textbf{1}\curlybrack{s_i\in [0,D]} \\
    &\geq -Dt.
\end{aligned}
\end{equation*}
We have shown that: $-Dt\leq \sum_{i=1}^ts_ig_i\leq 1$, and hence, 
\begin{equation}\label{eq:wealth_ub}
\abs{\sum_{i=1}^ts_ig_i}\leq \max \curlybrack{1,Dt}\leq Dt+1.    
\end{equation}
Next, we bound the distance between the consecutive predicted radii. Observe that for KT bettor:
    \begin{equation*}
    \begin{aligned}
        s_{t+1} &= -\frac{\sum_{i=1}^{t}g_i}{t+1} \roundbrack{1-\sum_{i=1}^{t}g_i s_i}\\
        &= -\frac{\sum_{i=1}^{t}g_i}{t+1} \roundbrack{1-\sum_{i=1}^{t-1}g_i s_i} + g_t s_t \frac{\sum_{i=1}^{t}g_i}{t+1}\\
        &= -\frac{\sum_{i=1}^{t-1}g_i}{t+1} \roundbrack{1-\sum_{i=1}^{t-1}g_i s_i}-\frac{g_t}{t+1} \roundbrack{1-\sum_{i=1}^{t-1}g_i s_i} + g_t s_t \frac{\sum_{i=1}^{t}g_i}{t+1}\\
        &= \frac{t}{t+1}s_t+\frac{1}{t+1} \roundbrack{-g_t+g_t\sum_{i=1}^{t-1}g_i s_i + g_t s_t \sum_{i=1}^{t}g_i},
    \end{aligned}
\end{equation*}
and hence,
\begin{equation}\label{eq:kt_bettor_iterates}
    s_{t+1} - s_{t} = \frac{1}{t+1} \roundbrack{-s_t-g_t+g_t\sum_{i=1}^{t-1}g_i s_i + g_t s_t \sum_{i=1}^{t}g_i}.
\end{equation}
From~\eqref{eq:kt_bettor_iterates} and~\eqref{eq:wealth_ub}, it follows that:
\begin{equation*}
    \abs{s_{t+1} - s_{t}} \leq \frac{1}{t+1} \roundbrack{D+1+D(t-1) +1+ Dt} \leq 2D+1.
\end{equation*}
Combining that with the fact that $s_1=0\in [0,D]$ and the result in step 1, we conclude that the iterates of the KT algorithm are bounded: $\abs{s_t}\leq 3D+1$.

\item Finally, we show that if~\eqref{eq:emp_coverage} fails to hold, then the iterates of KT bettor can not be bounded. Note that:
\begin{equation*}
    \abs{\frac{1}{t}\sum_{i=1}^t \indicator{Y_i \notin \hat{C}_i(X_i)} - \alpha} = \frac{1}{t}\abs{\sum_{i=1}^t g_i},
\end{equation*}
where $g_i$ are defined in Algorithm~\ref{alg:adapt_conf_kt}. Next, suppose that~\eqref{eq:emp_coverage} is not true, that is, $\exists \varepsilon>0: \forall T \ \exists T'>T: \frac{1}{T'}\abs{\sum_{i=1}^{T'} g_i}\geq \varepsilon$. Since
\begin{equation*}
    \abs{s_{t+1}} = \abs{\lambda_{t+1} W_{t}} = \frac{1}{t+1}\abs{\sum_{i=1}^{t} g_i} \cdot W_t,
\end{equation*}
we have that $\exists  \varepsilon>0: \forall T \ \exists T'>T$ such that:
\begin{equation*}
    \abs{s_{T'+1}} \geq \frac{1}{T'+1}\abs{\sum_{i=1}^{T'} g_i} \cdot W_{T'} \geq \frac{T'}{T'+1} \varepsilon \cdot W_{T'}.
\end{equation*}
For KT bettor, it holds that~\citep{orabona2016coin_bet}:
\begin{equation*}
    W_t \geq \frac{1}{K\sqrt{t}} \exp\roundbrack{\frac{t}{4}\roundbrack{\frac{1}{t}\sum_{i=1}^t g_i}^2},
\end{equation*}
where $K>0$ is a universal constant. Hence, we know that $\forall T\ \exists T'>T:$
\begin{equation*}
    \abs{s_{T'+1}} \geq \frac{T'}{T'+1}\frac{\varepsilon}{K\sqrt{T'}} \exp\roundbrack{\frac{T'}{4}\varepsilon^2},
\end{equation*}
implying that the iterates are unbounded. Hence, we have reached a contradiction with the conclusion of step 2, and thus, the coverage guarantee~\eqref{eq:emp_coverage} has to hold. This completes the proof.

\end{enumerate}

\end{proof}

\section{Additional Experiments}\label{appsec:add_exps}

In this Section, we present additional simulations to Section~\ref{sec:exps}. Section~\ref{appsubsec:changepoint_weighted_ls} is deferred to the changepoint setting. Section~\ref{appsubsec:electricity_demand} is deferred to the experiment with electricity demand dataset~\citep{harries1999elect}.

\subsection{Changepoint Setting and Weighted Least Squares Model}\label{appsubsec:changepoint_weighted_ls} 

\begin{figure}[!ht]
    \centering
    \includegraphics[width=0.8\linewidth]{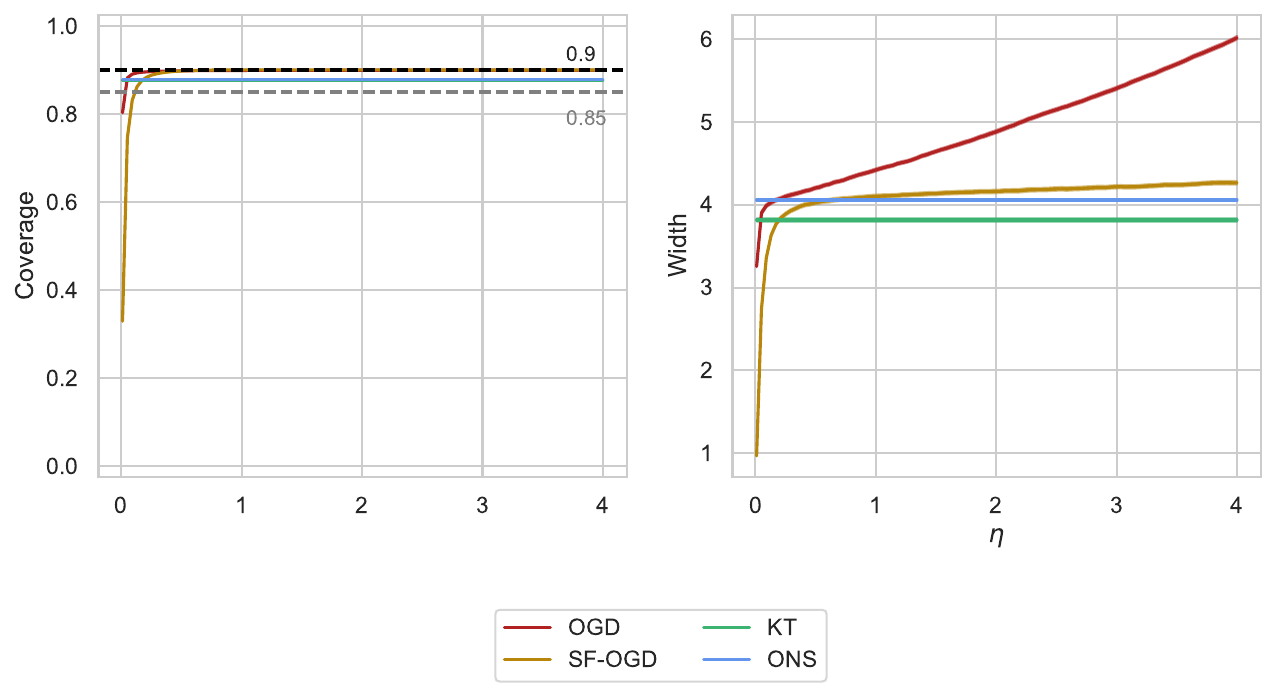}
    \caption{Comparison of our conformal predictor against those learned via OGD/SF-OGD with varying learning rates.  We observe that the performance of the proposed parameter-free approaches is close or matches that of the competitors with carefully tuned learning rates. Importantly, it avoids outputting overly conservative sets.}
    \label{fig:changepoint_summary_weighted_ls}
\end{figure}

Here, we compare adaptive conformal predictors that are learned using parameter-free optimization techniques against those that are trained via versions of online gradient descent, and hence, require specifying the learning rates (see Section~\ref{sec:methodology} for details). As an underlying model, we use a linear model, whose coefficients are learned by optimizing the weighted least squares objective:
\begin{equation*}
    \min_\beta \sum_{i=1}^t w_i (Y_i-X_i^\top \beta)^2.
\end{equation*}
Specifically, with $t$ available training points, the weights $(w_i)_{i=1}^t$ are assigned to the first $t$ (ordered) points, where $w_i = 0.99^{t+1-i}$, $i=1,\dots, t$. The results for varying learning rates are presented in Figure~\ref{fig:changepoint_summary_weighted_ls}. Similar to the case of a standard linear model, adaptive conformal predictors that utilize betting scheme tends to slightly undercover after processing 2000 observations. Using learning rates that are too high results in conformal predictors that output overly conservative sets. For example, OGD with $\eta=4$ yields conformal predictors that output sets which are more than 50\% larger than those corresponding to KT betting.

\begin{figure}[!ht]
    \centering
    \includegraphics[width=\linewidth]{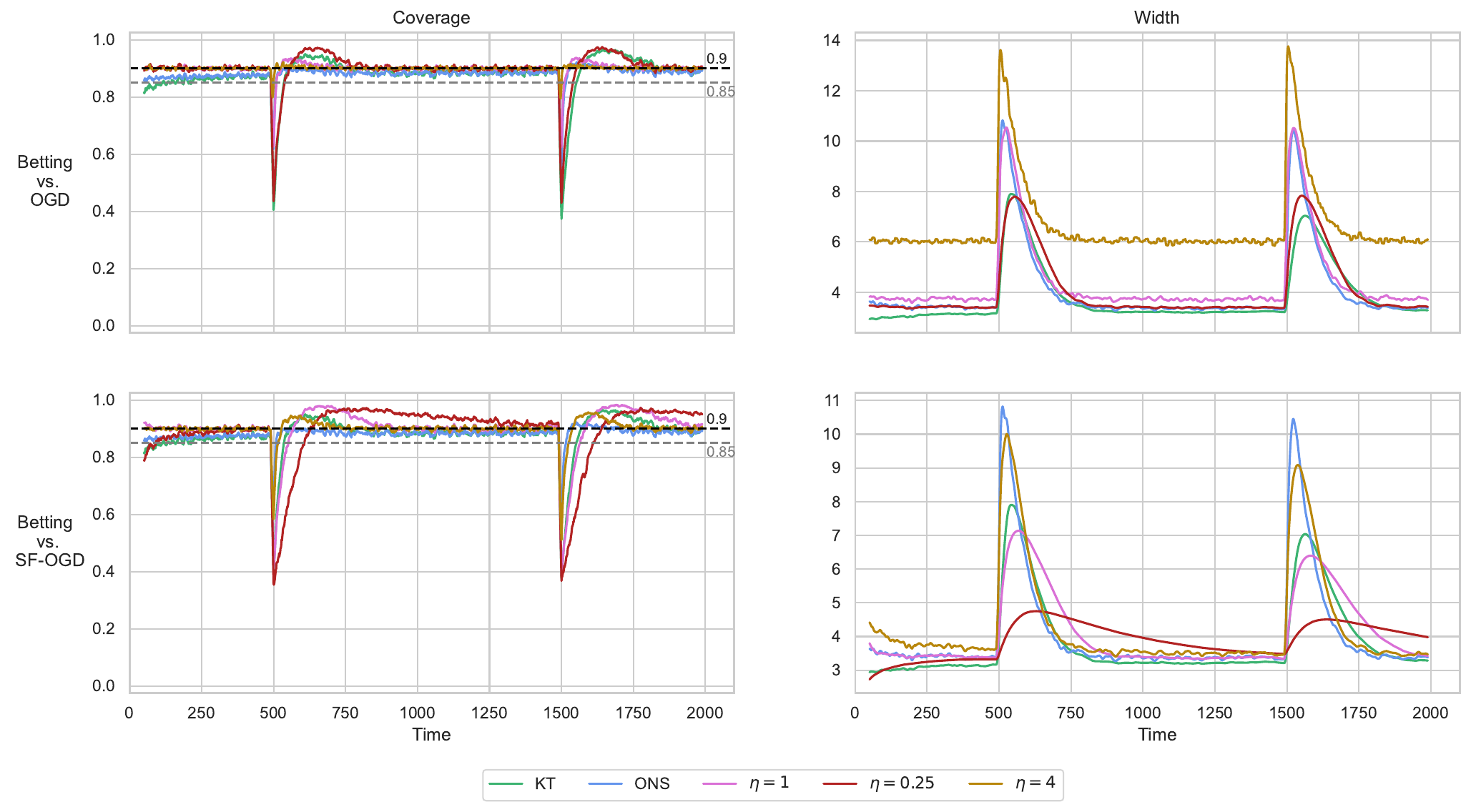}
    \caption{Performance of several methods when a linear model, whose coefficients are learned by optimizing the weighted least squares objective. We observe that the performance of the proposed parameter-free approaches is close or matches that of the competitors with carefully tuned learning rates. The results are aggregated over 250 random seeds and smoothed using rolling window of size 10.}
    \label{fig:ws_changepoint}
\end{figure}

The results localized coverage and width for a subset of learning rates are presented in Figure~\ref{fig:ws_changepoint}. We observe that the performance of the proposed parameter-free approaches is close or matches that of the competitors with carefully tuned learning rates. Our conformal predictor quickly restore coverage after a change in distribution has occurred and avoid being overly conservative once an underlying model adapts to the new settings (see bottom-left plot and $\eta=0.25$ or $\eta=1$).

\clearpage

\subsection{Electricity Demand Dataset}\label{appsubsec:electricity_demand}

\begin{figure}[!ht]
    \centering
    \subfigure[Coverage.]{\includegraphics[width=0.45\linewidth]{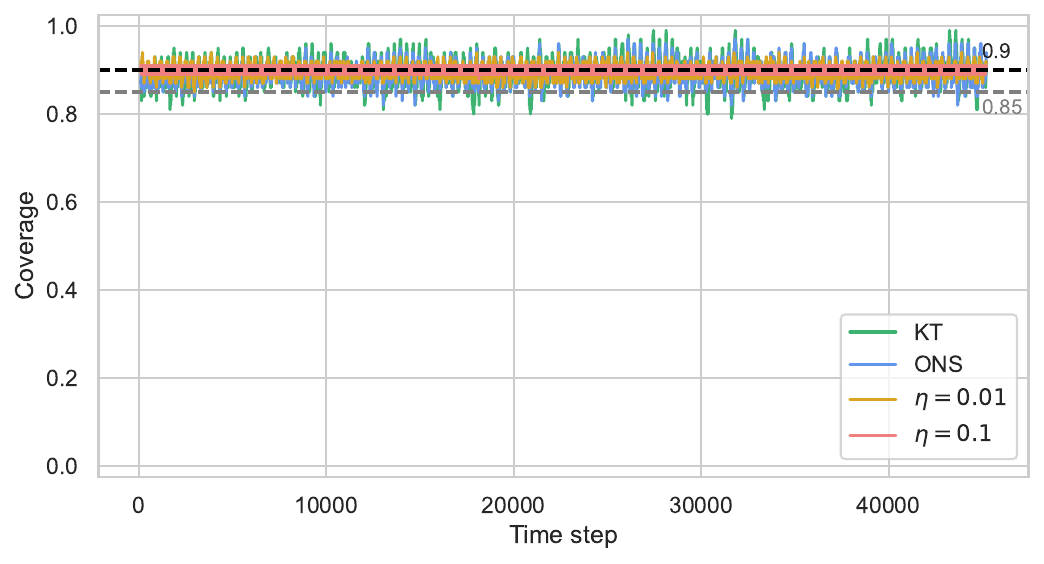}}
    \hspace{0.25in}
    \subfigure[Width.]{\includegraphics[width=0.45\linewidth]{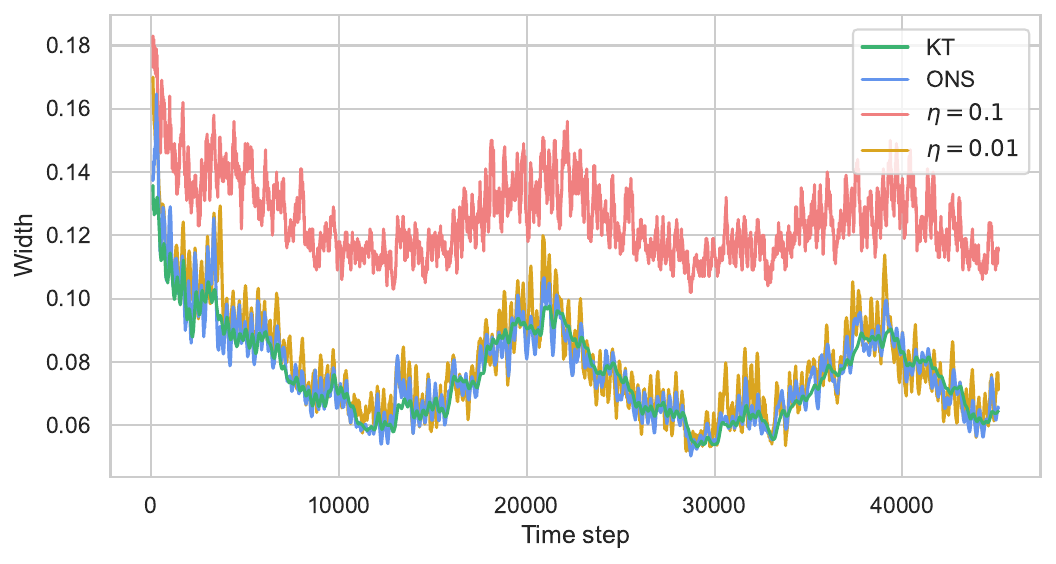}}
    \caption{Comparison between conformal predictors constructed using betting schemes and versions of OGD~\eqref{eq:ogd_update} with learning rates $\eta\in\{0.01,0.1\}$. For all methods the empirical coverage is near the nominal level. The width of a conformal predictor based on OGD with learning rate $\eta = 0.1$ is consistently higher than that of other methods. The results are smoothed over a rolling window of 100 observations.}
    \label{fig:electricity_ogd}
\end{figure}

In Figure~\ref{fig:electricity_ogd}, we compare coverage and width (smoothed over a rolling window of 100 observations) of conformal predictors constructed using betting schemes against those based on OGD with varying learning rates. While for all methods the empirical coverage is near the nominal level, the width of a conformal predictor based on OGD with learning rate $\eta = 0.1$ is consistently higher than that of other methods.

\begin{figure}[!ht]
    \centering
    \subfigure[Higher learning rate.]{\includegraphics[width=0.4\linewidth]{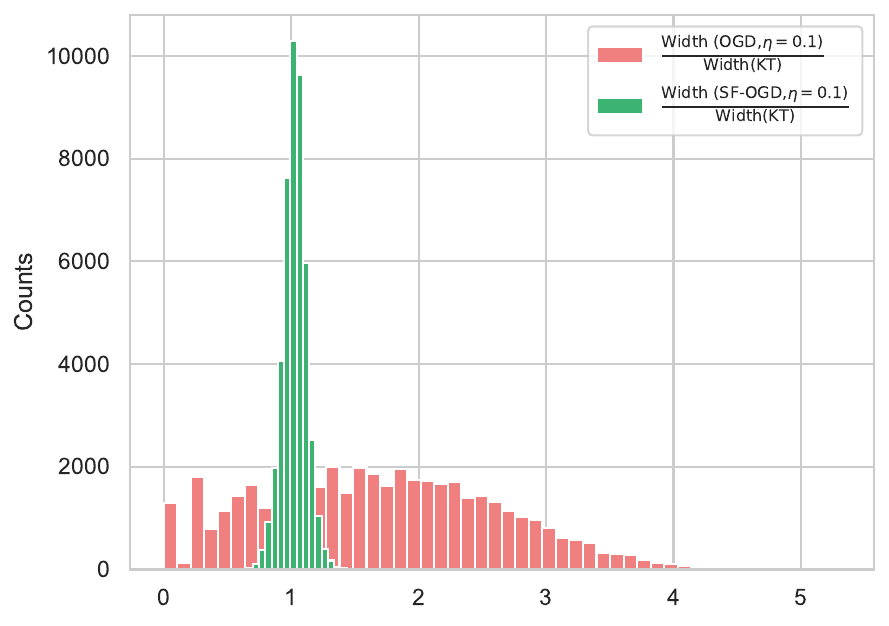}}
    \hspace{0.25in}
    \subfigure[Lower learning rate.]{\includegraphics[width=0.4\linewidth]{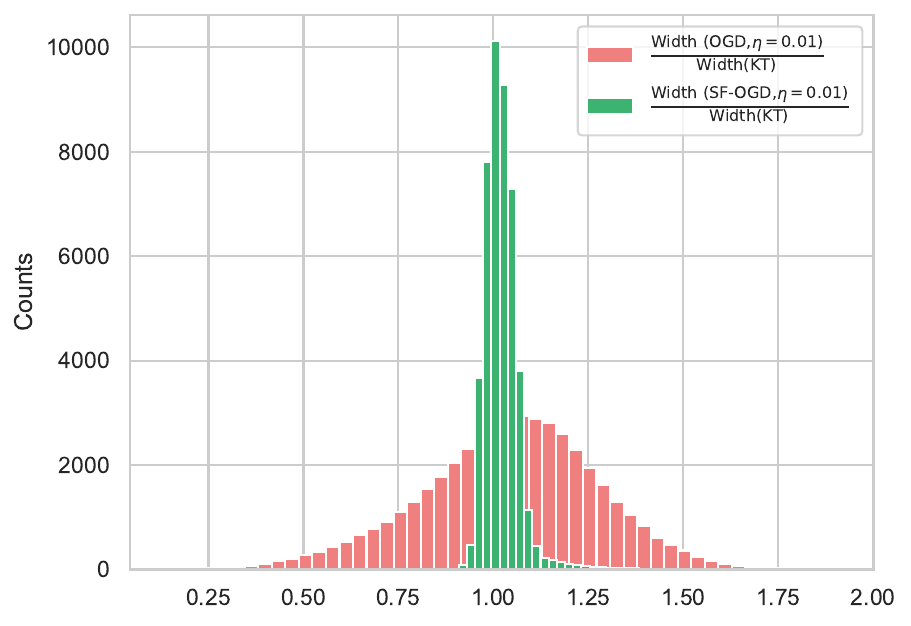}}
    \caption{The histograms for the ratios of the widths of the prediction intervals obtained from conformal predictors based on versions of online gradient descent to that of conformal predictors based on KT betting scheme. The average width of intervals corresponding to OGD with learning rate $\eta=0.1$ is almost 70\% larger than that of KT-based conformal predictor, whereas for SF-OGD with the same learning rate the number reduces to 3\%. For lower learning rates, the average width are almost equal.}
    \label{fig:electricity_histograms}
\end{figure}

In Figure~\ref{fig:electricity_histograms}, we demonstrate the histograms for the ratios of the widths of the prediction intervals obtained from conformal predictors based on versions of online gradient descent to that of conformal predictors based on KT betting scheme. Ignoring the first 100 observations (warm-up period), the average width of intervals corresponding to OGD with learning rate $\eta=0.1$ is almost 70\% larger than that of KT-based conformal predictor. For SF-OGD with the same learning rate, this number reduces to only 3\%. 

\begin{figure}[!ht]
    \centering
    \includegraphics[width=0.95\linewidth]{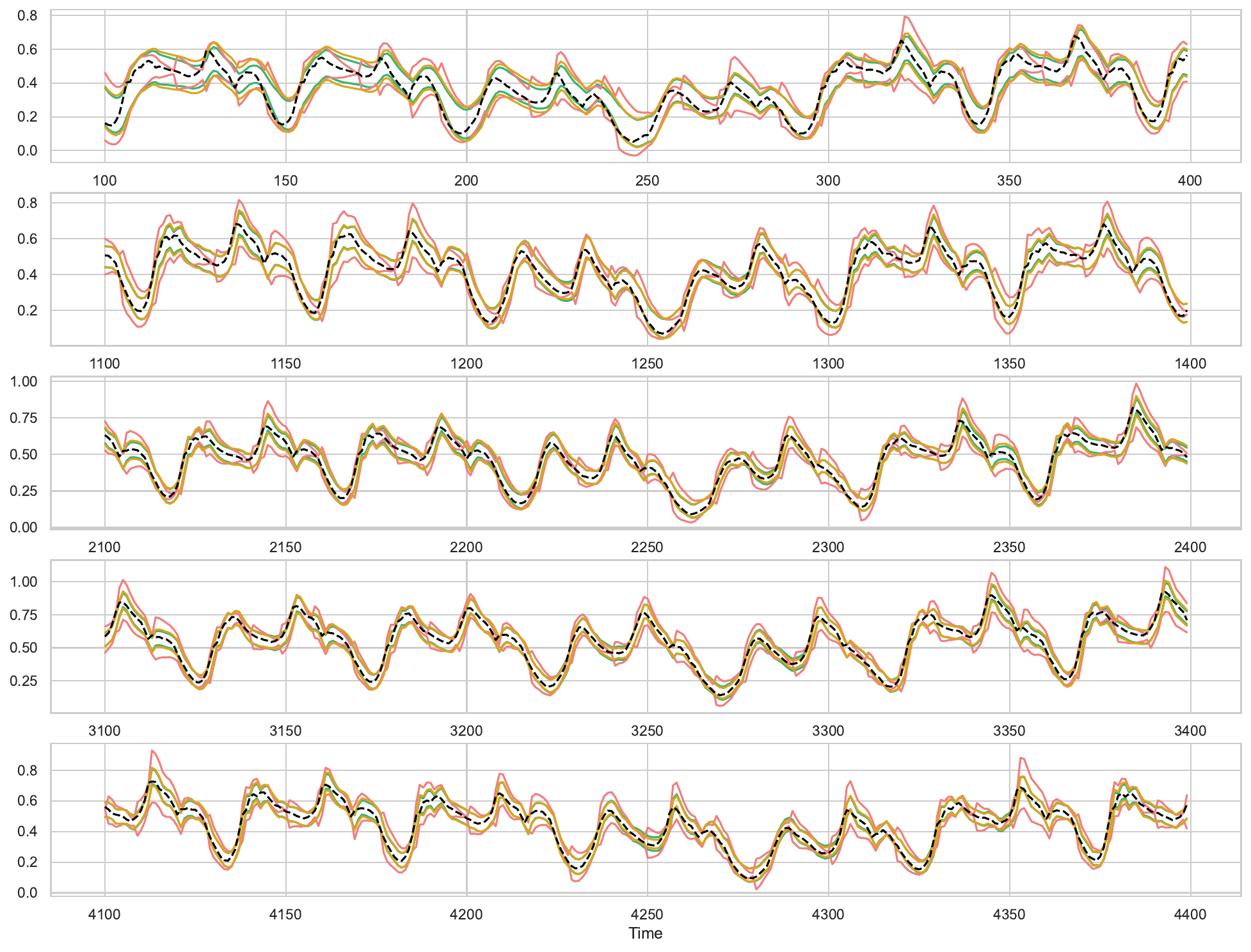}
    \caption{Prediction bands for conformal predictors which are learned via KT-betting (green), OGD with learning rate $\eta=0.01$ (yellow), and OGD with learning rate $\eta=0.1$ (coral). Learning rate $\eta=0.1$ yields conformal predictors that output overly large prediction sets across all time steps. KT-based and learning rate $\eta=0.01$ bands are visually very close, particularly for later time steps.}
    \label{fig:kt_ogd_ts}
\end{figure}

In Figure~\ref{fig:kt_ogd_ts}, we compare the KT-based conformal predictor against that based on OGD with either of two learning rates: 0.01 or 0.1. The prediction bands for conformal predictors based on KT-betting and OGD with learning rate $\eta=0.01$ are visually very close, particularly for later time steps. Conformal predictor based on OGD with learning rate $\eta=0.1$ yields sets that are generally larger.

\begin{figure}[!ht]
    \centering
    \includegraphics[width=0.95\linewidth]{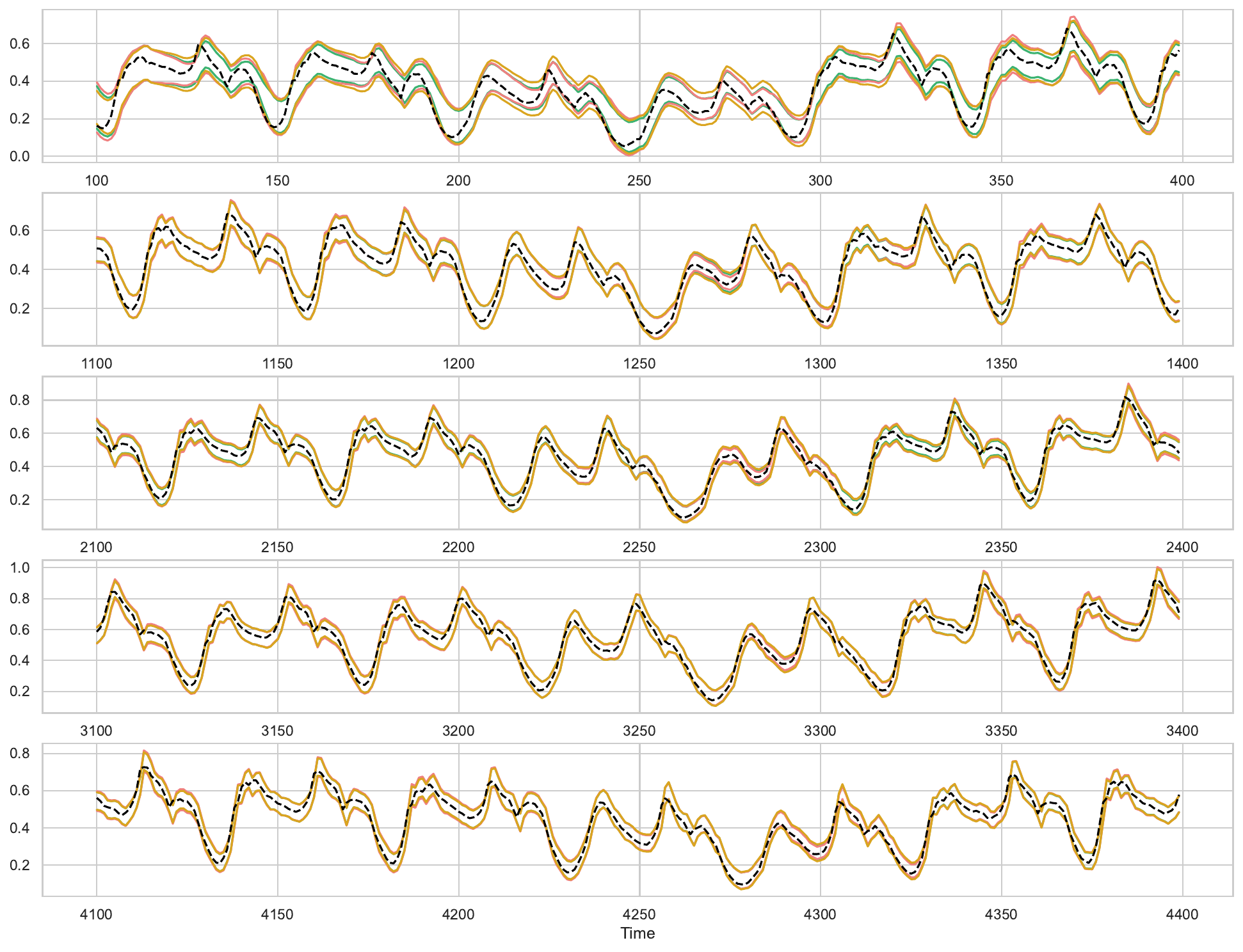}
    \caption{Prediction bands for conformal predictors which are learned via KT-betting (green), SF-OGD with learning rate $\eta=0.01$ (yellow), and SF-OGD with learning rate $\eta=0.1$ (coral). While for the initial time steps the bands are close, they become visually indistinguishable for larger time steps. The difference between two learning rates essentially disappears due to effective learning rate that decreases over time.}
    \label{fig:kt_sfogd_ts}
\end{figure}

In Figure~\ref{fig:kt_sfogd_ts}, we compare KT-based conformal predictor against that based on SF-OGD with the same learning rates: 0.01 or 0.1. The prediction bands for conformal predictors based on KT-betting and SF-OGD with either of the learning rates become visually indistinguishable, especially for later time steps. The difference between the outputs of conformal predictors based on SF-OGD with different learning rates diminishes due to effective learning rate that decreases over time.

\clearpage

\subsection{Stock Prices Data}\label{appsubsec:stocks}

\begin{figure}[!ht]
    \centering
    \subfigure[1-day ahead.]{\includegraphics[width=0.45\linewidth]{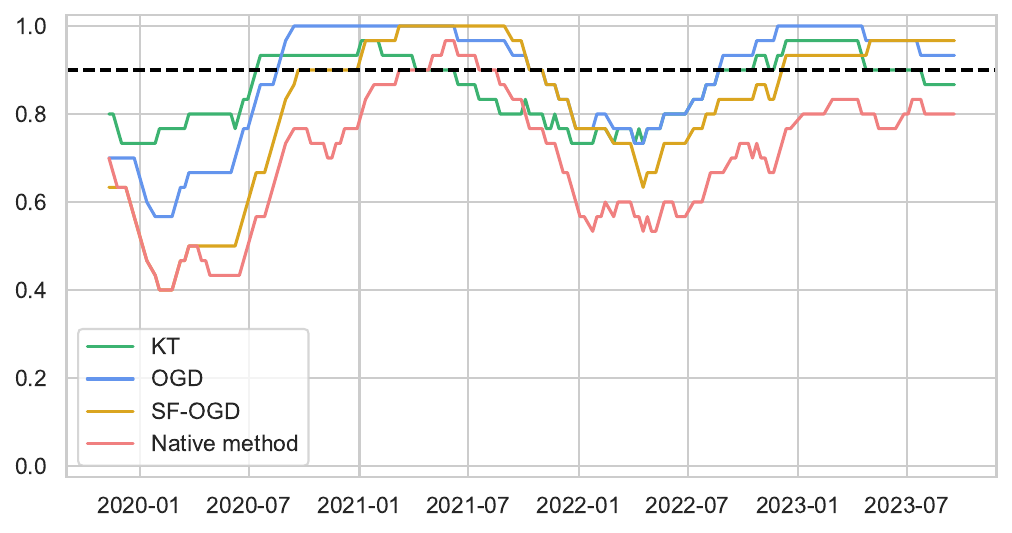}}
    \hspace{0.25in}
    \subfigure[5-days ahead.]{\includegraphics[width=0.45\linewidth]{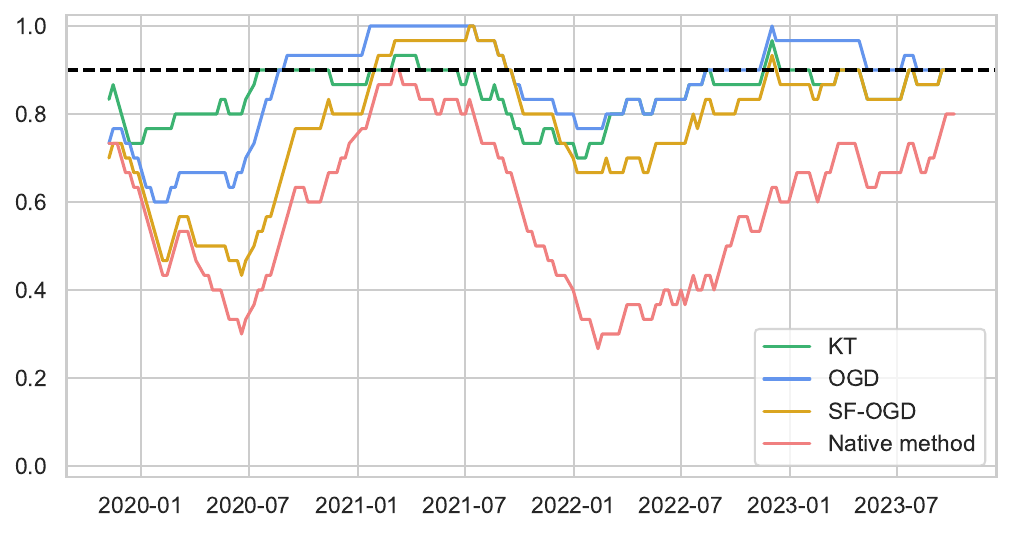}}
    \subfigure[1-day ahead.]{\includegraphics[width=0.45\linewidth]{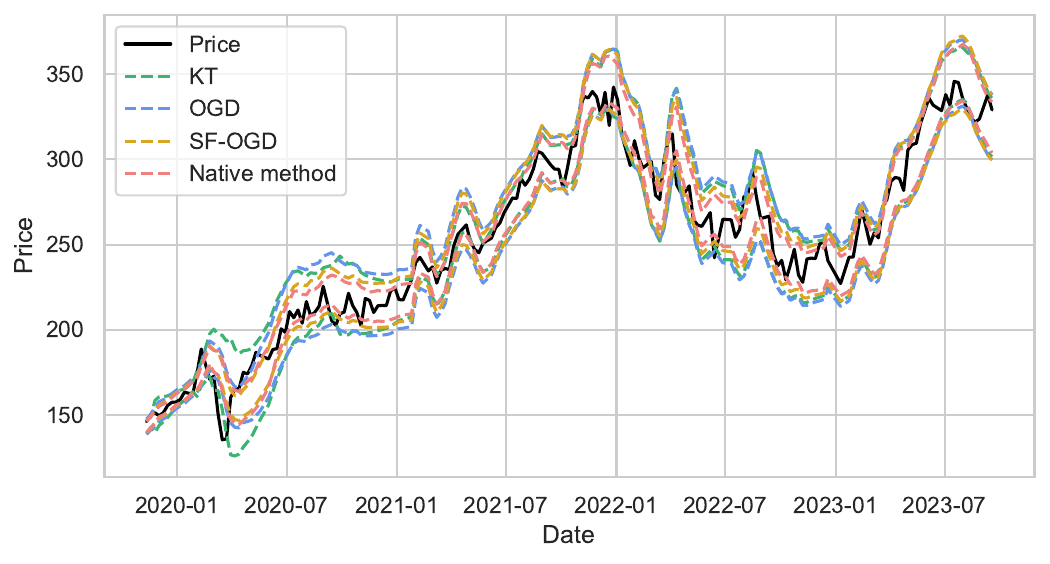}}
    \hspace{0.25in}
    \subfigure[5-days ahead.]{\includegraphics[width=0.45\linewidth]{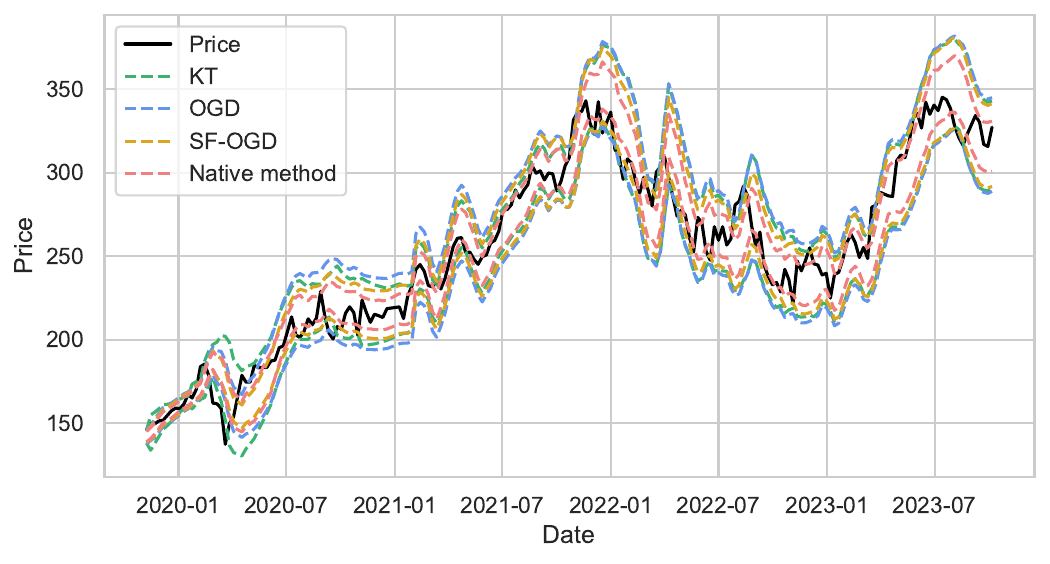}}
    \caption{Top row: localized coverage for \texttt{MSFT} stock. The results are averaged over a rolling window of size 30. Bottom row: stock prices on Fridays plotted along with prediction bands corresponding to different methods.}
    \label{fig:msft_stock}
\end{figure}

\begin{figure}[!ht]
    \centering
    \subfigure[1-day ahead.]{\includegraphics[width=0.45\linewidth]{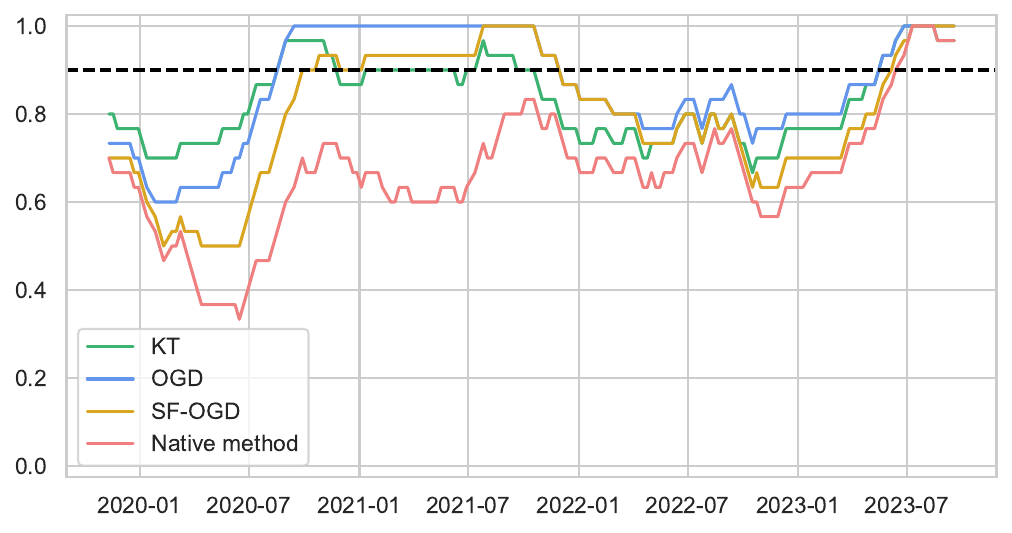}}
    \hspace{0.25in}
    \subfigure[5-days ahead.]{\includegraphics[width=0.45\linewidth]{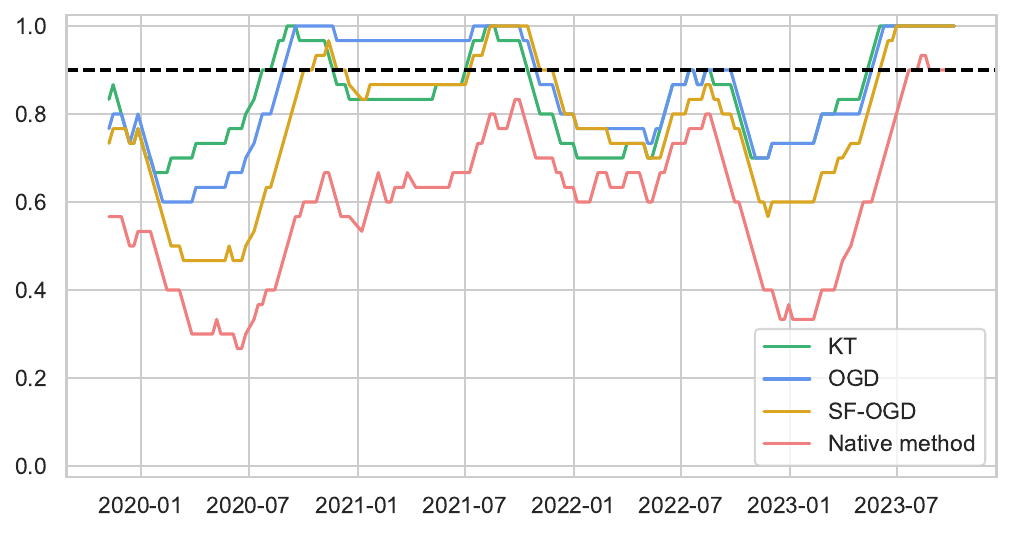}}
    \subfigure[1-day ahead.]{\includegraphics[width=0.45\linewidth]{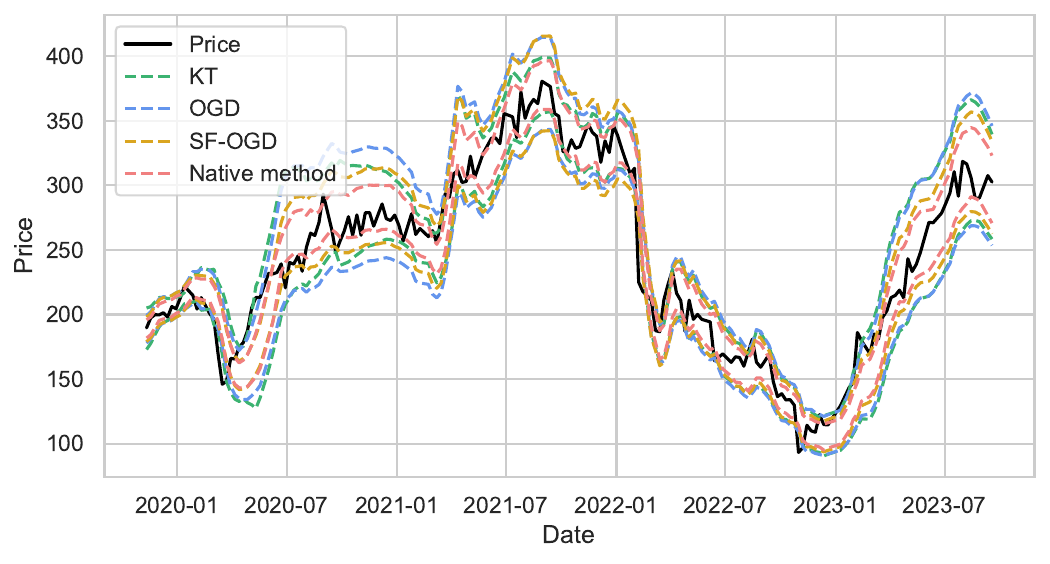}}
    \hspace{0.25in}
    \subfigure[5-days ahead.]{\includegraphics[width=0.45\linewidth]{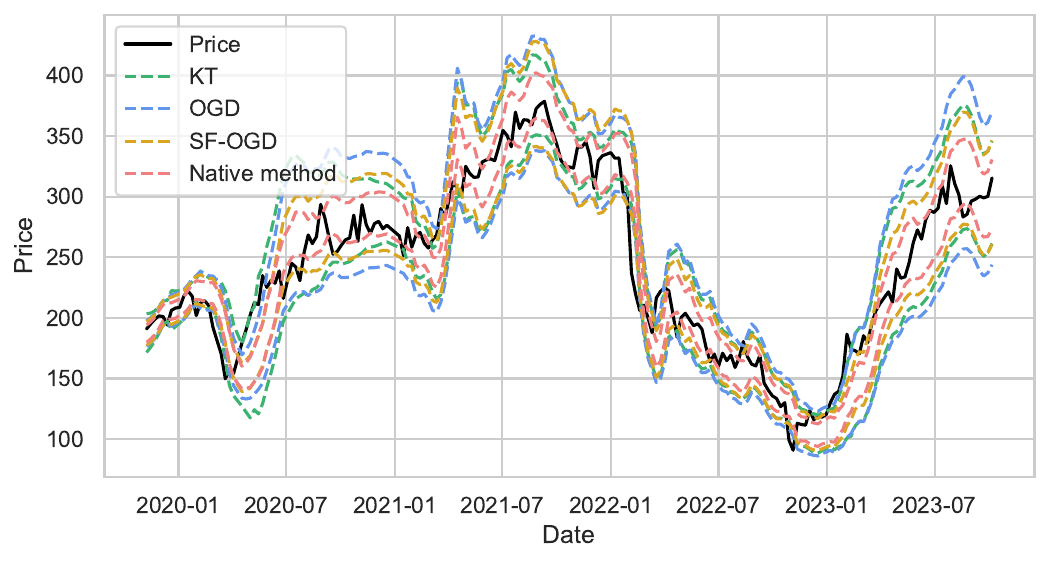}}
    \caption{Top row: localized coverage for \texttt{META} stock. The results are averaged over a rolling window of size 30. Bottom row: stock prices on Fridays plotted along with prediction bands corresponding to different methods.}
    \label{fig:meta_stock}
\end{figure}

\begin{figure}[!ht]
    \centering
    \subfigure[1-day ahead.]{\includegraphics[width=0.45\linewidth]{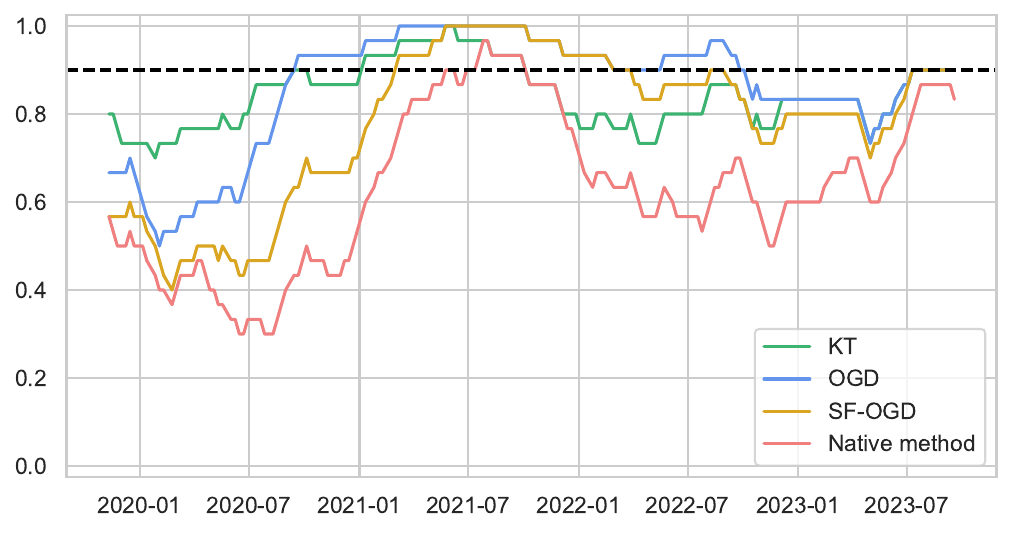}}
    \hspace{0.25in}
    \subfigure[5-days ahead.]{\includegraphics[width=0.45\linewidth]{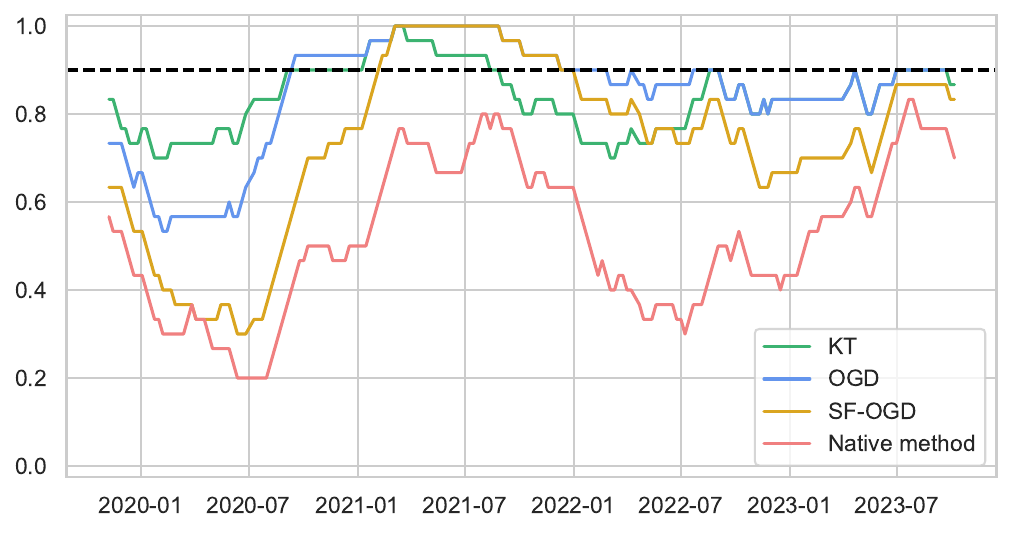}}
    \subfigure[1-day ahead.]{\includegraphics[width=0.45\linewidth]{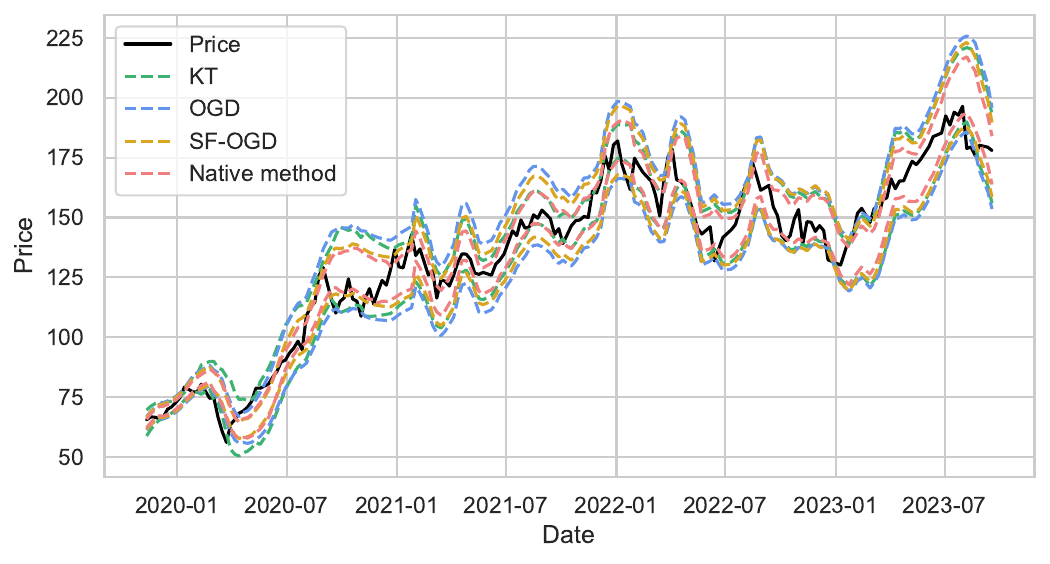}}
    \hspace{0.25in}
    \subfigure[5-days ahead.]{\includegraphics[width=0.45\linewidth]{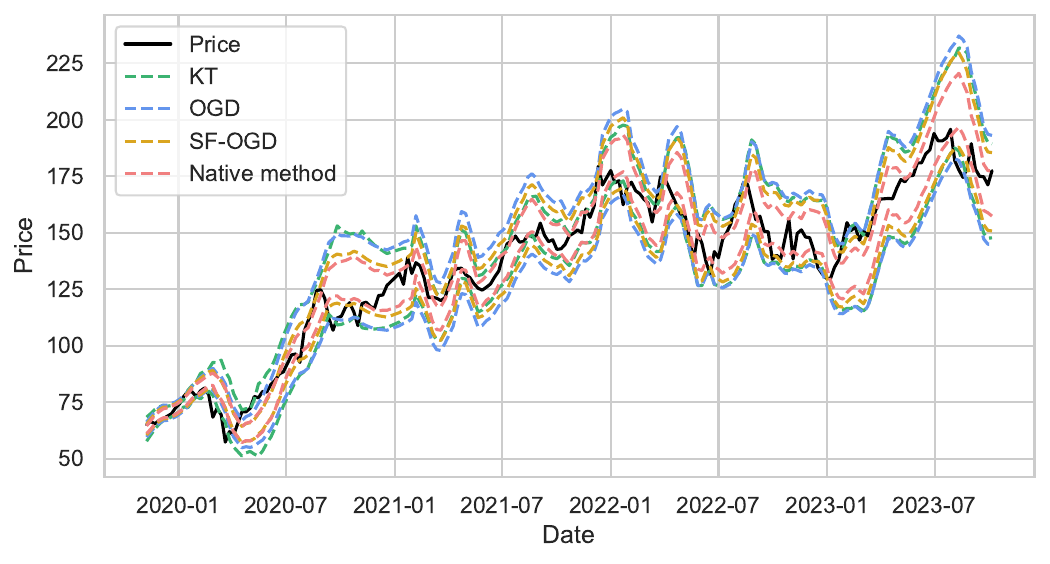}}
    \caption{Top row: localized coverage for \texttt{APPL} stock. The results are averaged over a rolling window of size 30. Bottom row: stock prices on Fridays plotted along with prediction bands corresponding to different methods.}
    \label{fig:apple_stock}
\end{figure}

\begin{figure}[!ht]
    \centering
    \subfigure[1-day ahead.]{\includegraphics[width=0.45\linewidth]{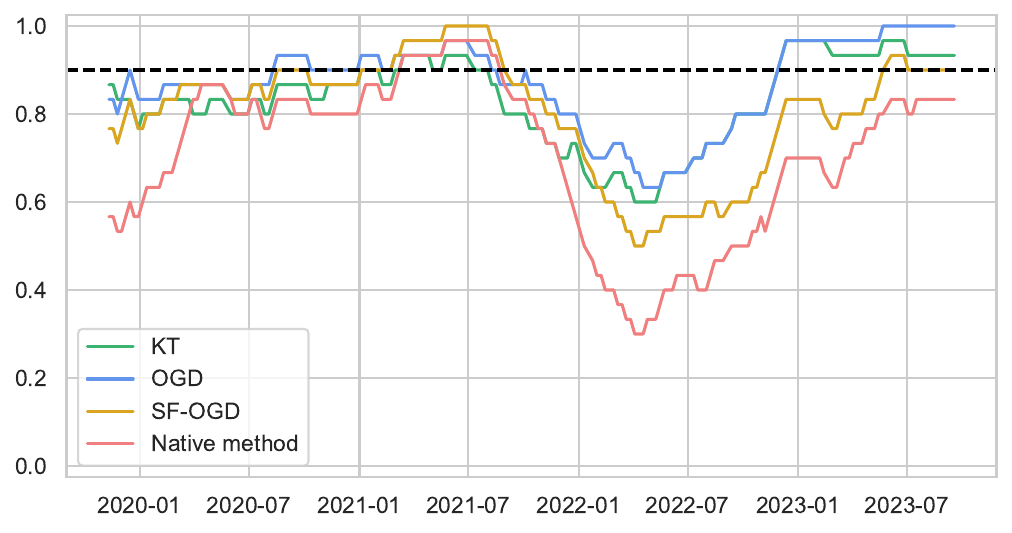}}
    \hspace{0.25in}
    \subfigure[5-days ahead.]{\includegraphics[width=0.45\linewidth]{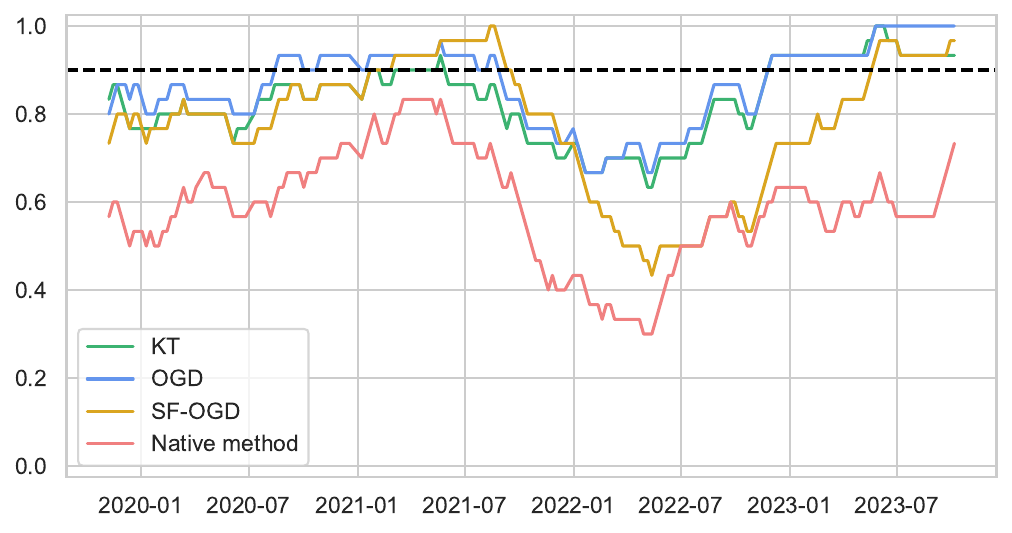}}
    \subfigure[1-day ahead.]{\includegraphics[width=0.45\linewidth]{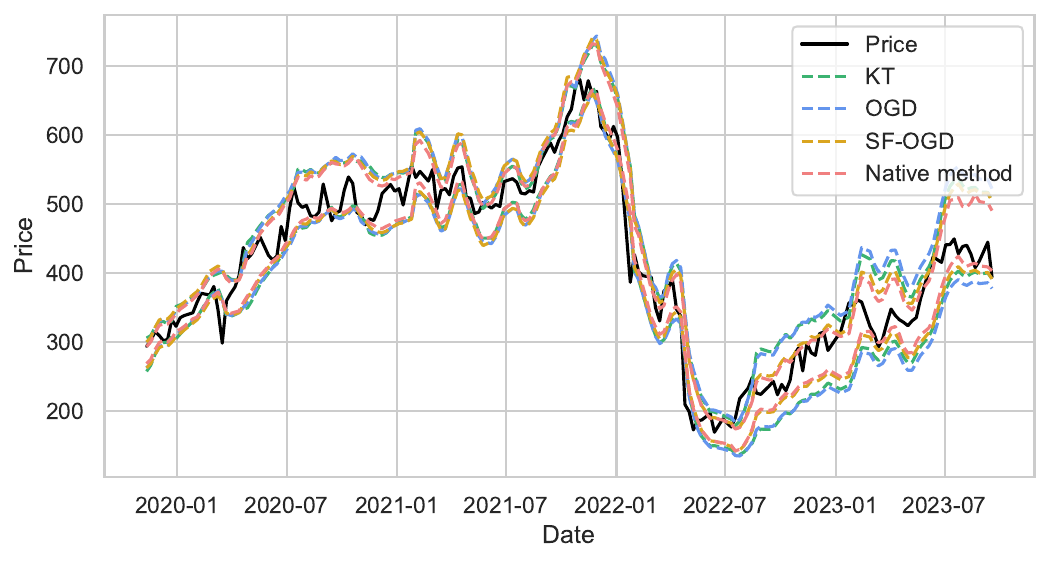}}
    \hspace{0.25in}
    \subfigure[5-days ahead.]{\includegraphics[width=0.45\linewidth]{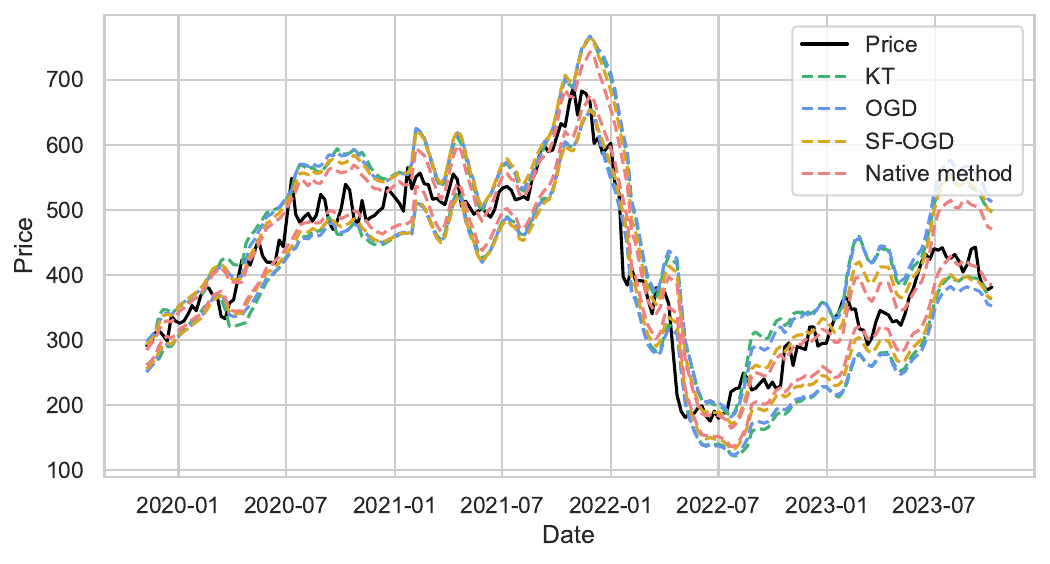}}
    \caption{Top row: localized coverage for \texttt{NFLX} stock. The results are averaged over a rolling window of size 30. Bottom row: stock prices on Fridays plotted along with prediction bands corresponding to different methods.}
    \label{fig:nflx_stock}
\end{figure}

\begin{figure}[!ht]
    \centering
    \subfigure[1-day ahead.]{\includegraphics[width=0.45\linewidth]{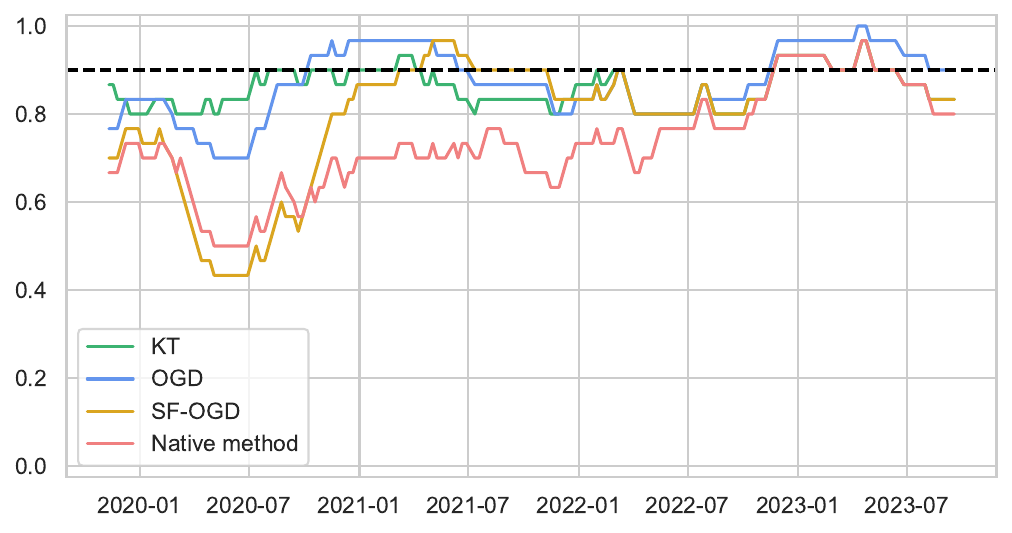}}
    \hspace{0.25in}
    \subfigure[5-days ahead.]{\includegraphics[width=0.45\linewidth]{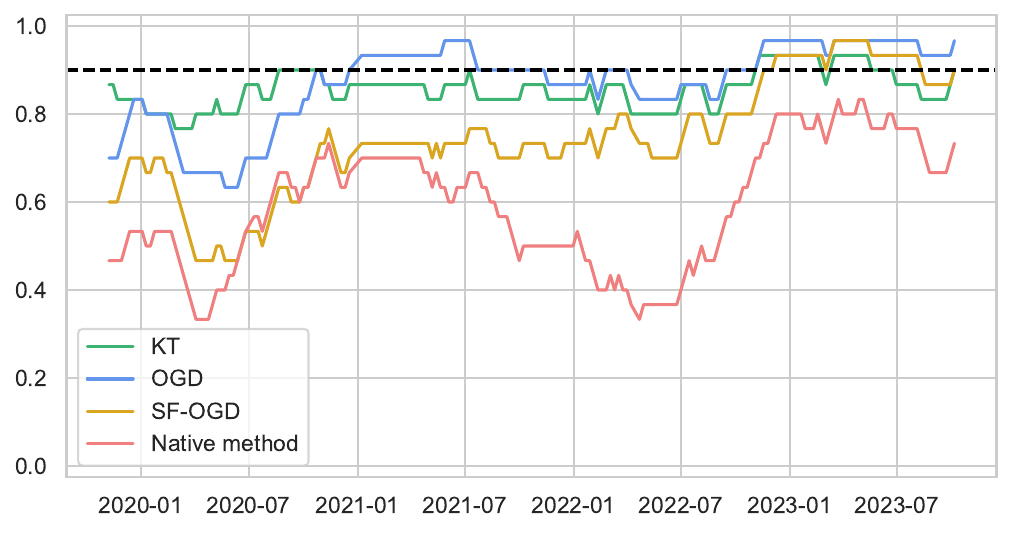}}
    \subfigure[1-day ahead.]{\includegraphics[width=0.45\linewidth]{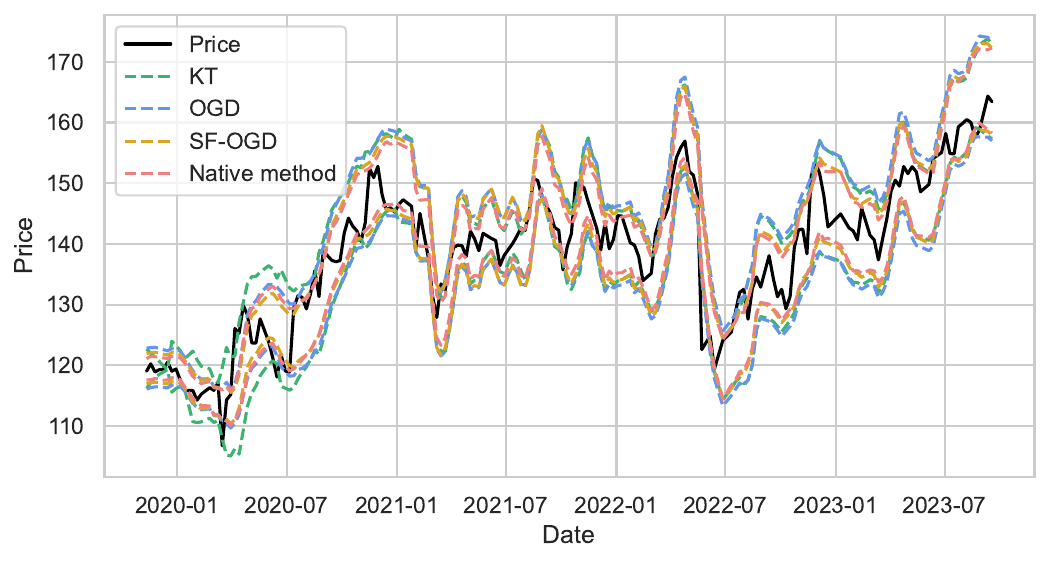}}
    \hspace{0.25in}
    \subfigure[5-days ahead.]{\includegraphics[width=0.45\linewidth]{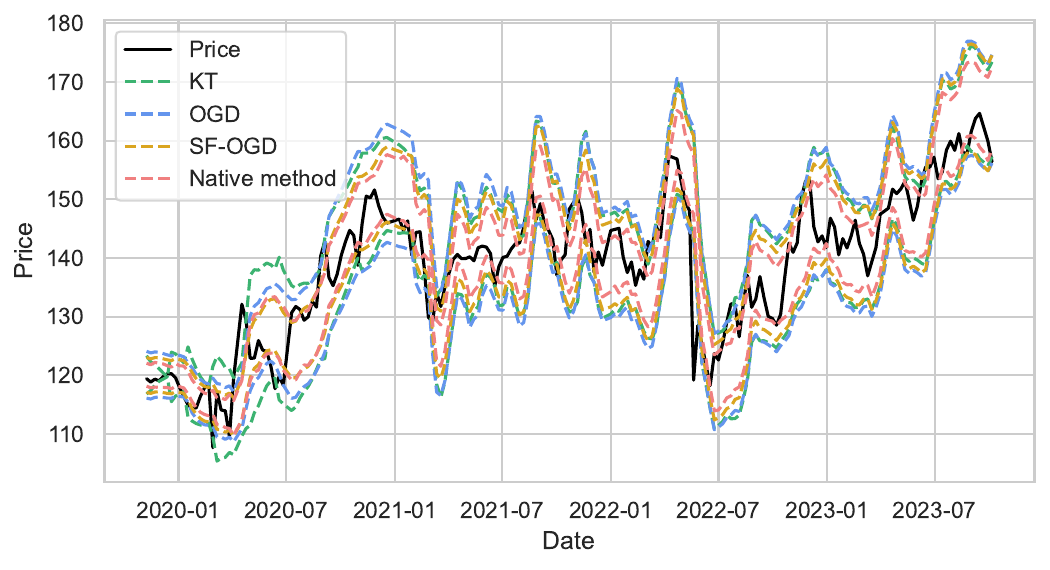}}
    \caption{Top row: localized coverage for \texttt{WMT} stock. The results are averaged over a rolling window of size 30. Bottom row: stock prices on Fridays plotted along with prediction bands corresponding to different methods.}
    \label{fig:wmt_stock}
\end{figure}

In this Section, we demonstrate the results of running different approaches for quantifying predictive uncertainty in stock price forecasting for \texttt{MSFT} (Figure~\ref{fig:msft_stock} and Table~\ref{table:msft_stock}), \texttt{META} (Figure~\ref{fig:meta_stock} and Table~\ref{table:meta_stock}), \texttt{APPL} (Figure~\ref{fig:apple_stock} and Table~\ref{table:apple_stock}), \texttt{NFLX} (Figure~\ref{fig:nflx_stock} and Table~\ref{table:nflx_stock}), and \texttt{WMT} (Figure~\ref{fig:wmt_stock} and Table~\ref{table:wmt_stock}).

\begin{table}[!ht]
\centering
\begin{tabular}{c|c c c c |c c c c} 
 \multicolumn{5}{c}{\hspace{0.1in} Coverage} & \multicolumn{4}{c}{Width} \\
 \hline
 $k$ & KT& OGD& SF-OGD& Native&KT& OGD& SF-OGD& Native\\
\hline
1 & 85.7 & 87.1 & 81.9 & 71.9 &32.2 &33.5 &27.8&21.1\\
 2 & 85.5 & 87.2 & 80.9 & 71.1&32.8 & 34.3& 31& 21.4\\
 3 & 85.5 & 86.8 & 80.3 & 63.7 &36.2 & 39.3 & 31.9 &  21.7\\
 4 & 85.2 & 86.5 & 77.8 & 63.5 & 38.2 & 42 & 34.6 & 21.4\\
 5 & 85 & 85.9 & 78 & 61.2 & 39.4 & 42.3 & 33.3 & 21.5
\end{tabular}
    \caption{The empirical coverage and average width of prediction intervals for \texttt{MSFT} stock.}
\label{table:msft_stock}
\end{table}

\clearpage

\begin{table}[!ht]
\centering
\begin{tabular}{c|c c c c |c c c c} 
 \multicolumn{5}{c}{\hspace{0.1in} Coverage} & \multicolumn{4}{c}{Width} \\
 \hline
 $k$ & KT& OGD& SF-OGD& Native&KT& OGD& SF-OGD& Native\\
\hline
1 & 83.8 & 86.2 & 80.5& 69 & 54.9& 60.7& 47.9& 31.6\\
 2 & 84.3 & 86.4 & 80 & 66 & 52.5& 64.1 & 50.2& 31.9\\
 3 & 84.6 & 86.8 & 79.1 & 61.5 & 56.7 & 68 & 55.5& 32.3\\
 4 & 84.3 & 85.7 & 78.3 & 58.7 & 63.6 & 74.1 & 55.3& 31.9\\
 5 & 84.6 & 85 & 78.4 & 60.8 & 63.5 & 76.4 & 55.5 & 32.1
\end{tabular}
    \caption{The empirical coverage and average width of prediction intervals for \texttt{META} stock.}
\label{table:meta_stock}
\end{table}

\clearpage

\begin{table}[!ht]
\centering
\begin{tabular}{c|c c c c |c c c c} 
 \multicolumn{5}{c}{\hspace{0.1in} Coverage} & \multicolumn{4}{c}{Width} \\
 \hline
 $k$ & KT& OGD& SF-OGD& Native&KT& OGD& SF-OGD& Native\\
\hline
1 & 84.8 & 84.8 & 77.1 & 64.3 &21.8 & 27.1 & 20.9& 14\\
 2 & 84.7 & 85.5 & 77 & 63 &24.3 & 25.8& 22.1& 14.1\\
 3 &84.6& 85.5& 76.1& 58.5 & 23.6 & 28.7 & 22.5& 14.3\\
 4 & 84.3& 85.2& 74.3& 54.8 & 24.5 & 28.3& 22.4& 14.1\\
 5 & 84.1 & 84.1 & 73.6 & 53.3 & 27.5 & 31.7 & 23.2 & 14.2
\end{tabular}
    \caption{The empirical coverage and average width of prediction intervals for \texttt{APPL} stock.}
\label{table:apple_stock}
\end{table}

\clearpage

\begin{table}[!ht]
\centering
\begin{tabular}{c|c c c c |c c c c} 
 \multicolumn{5}{c}{\hspace{0.1in} Coverage} & \multicolumn{4}{c}{Width} \\
 \hline
 $k$ & KT& OGD& SF-OGD& Native&KT& OGD& SF-OGD& Native\\
\hline
1 & 83.8 & 86.7 & 80.5 & 70.5 &78.7& 84.6& 68.8 & 55.3\\
 2 & 84.3 & 86.8 & 80.9 & 69.8 & 79 &  87.5 & 69.7 & 55.3\\
 3 &83.3 & 86.8 & 80.3 & 64.1 & 82.9&  94.8&  76.8 & 55.8\\
 4 & 83.5& 87 & 79.1 & 62.6 & 88.9& 96.5& 84.5 & 55.5\\
 5 & 83.7& 86.3& 78 & 59.9 & 105.2 & 105.6& 85.9 & 55.4
\end{tabular}
    \caption{The empirical coverage and average width of prediction intervals for \texttt{NFLX} stock.}
\label{table:nflx_stock}
\end{table}

\clearpage

\begin{table}[!ht]
\centering
\begin{tabular}{c|c c c c |c c c c} 
 \multicolumn{5}{c}{\hspace{0.1in} Coverage} & \multicolumn{4}{c}{Width} \\
 \hline
 $k$ & KT& OGD& SF-OGD& Native&KT& OGD& SF-OGD& Native\\
\hline
1 & 85.7& 85.2 & 78.1 & 71.4 & 13.5 & 12.7 & 10.4 & 9.1\\
 2 &86 & 86 & 78.7 & 68.1 & 13.4 & 13.1 & 11.8 & 9.2\\
 3 & 86.3 & 86.3& 78.2 & 66.7 & 14.6 & 14.3 & 11.7 & 9.2\\
 4 & 85.7 & 85.2& 75.2& 62.6 &15.2& 15.4& 12.4& 9.2\\
 5 & 85.5 & 85.5 & 73.6 & 58.6 & 16.9& 16.5& 12.9& 9.2
\end{tabular}
    \caption{The empirical coverage and average width of prediction intervals for \texttt{WMT} stock.}
\label{table:wmt_stock}
\end{table}

\end{document}